%% file: main.tex
\documentclass[10pt,twocolumn,letterpaper]{article}

\usepackage{iccv}
\usepackage{times}
\usepackage{epsfig}
\usepackage{graphicx}
\usepackage{amsmath}
\usepackage{amssymb}

\usepackage{lipsum}
\usepackage{amsthm}
\usepackage[dvipsnames]{xcolor}
\usepackage{gensymb}
\usepackage{comment}
\usepackage{multirow}
\usepackage{subcaption}
\usepackage{booktabs}
\usepackage{microtype}
\usepackage{xspace}
\usepackage{enumitem}
\usepackage[bottom]{footmisc}
\usepackage{nimbusmononarrow}
\usepackage[linesnumbered,ruled]{algorithm2e}
\usepackage{algpseudocode}

\usepackage[pagebackref=true,breaklinks=true,letterpaper=true,colorlinks,bookmarks=false]{hyperref}

\hypersetup{
    colorlinks,
    citecolor=CornflowerBlue,
}

\input{notation}

\newcommand{\barf}{\color{black}}

\iccvfinalcopy 

\def\iccvPaperID{****} 

\begin{document}

\title{BARF \includegraphics[scale=0.08]{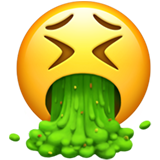}: {\barf B}undle-{\barf A}djusting Neural {\barf R}adiance {\barf F}ields}


\author{
  Chen-Hsuan Lin\textsuperscript{1} \quad
  Wei-Chiu Ma\textsuperscript{2} \quad
  Antonio Torralba\textsuperscript{2} \quad
  Simon Lucey\textsuperscript{1,3} \vspace{2pt} \\
  \textsuperscript{1}Carnegie Mellon University \quad
  \textsuperscript{2}Massachusetts Institute of Technology \quad
  \textsuperscript{3}The University of Adelaide \vspace{2pt} \\
  {\fontsize{10}{10}\selectfont \url{https://chenhsuanlin.bitbucket.io/bundle-adjusting-NeRF}}
}

\maketitle

\begin{abstract}
\input{0-abstract}
\end{abstract}

\input{1-introduction}
\input{2-relatedwork}
\input{3-approach}
\input{4-experiments}
\input{5-conclusion}


\renewcommand{\thesection}{\Alph{section}}
\setcounter{section}{0}
\input{A-supplementary}

{\small
\bibliographystyle{ieee_fullname}
\bibliography{reference}
}

\end{document}

%% file: notation.tex
\newcommand{\eucnorm}[1]{\left\|{#1}\right\|_2}

\newcommand{\I}{\mathcal{I}}
\newcommand{\p}{\mathbf{p}}
\newcommand{\x}{\mathbf{x}}
\newcommand{\0}{\mathbf{0}}
\newcommand{\W}{\mathcal{W}}
\newcommand{\delp}{\Delta\p}

\newcommand{\bTheta}{\boldsymbol{\Theta}}
\renewcommand{\u}{\mathbf{u}}
\newcommand{\uhom}{\bar{\u}}
\newcommand{\Ihat}{\hat{\I}}
\renewcommand{\c}{\mathbf{c}}
\newcommand{\y}{\mathbf{y}}

\newcommand{\Rot}{\mathbf{R}}
\newcommand{\trans}{\mathbf{t}}

\renewcommand{\o}{\mathbf{o}}
\newcommand{\X}{\mathbf{X}}
\newcommand{\1}{\mathbf{1}}
\newcommand{\U}{\mathbf{U}}
\renewcommand{\S}{\mathbf{S}}
\newcommand{\V}{\mathbf{V}}

\newcommand{\J}{\mathbf{J}}
\newcommand{\A}{\mathbf{A}}

\newcommand{\pderiv}[2]{\frac{\partial #1}{\partial #2}}

\newcommand{\SFM}{S\textit{f}M\xspace}

\newcommand{\eye}{\mathbf{I}}
\newcommand{\Real}{\mathbb{R}}

%% file: 0-abstract.tex

Neural Radiance Fields (NeRF)~\cite{mildenhall2020nerf} have recently gained a surge of interest within the computer vision community for its power to synthesize photorealistic novel views of real-world scenes.
One limitation of NeRF, however, is its requirement of accurate camera poses to learn the scene representations.
In this paper, we propose Bundle-Adjusting Neural Radiance Fields (BARF) for training NeRF from imperfect (or even unknown) camera poses --- the joint problem of learning neural 3D representations and registering camera frames.
We establish a theoretical connection to classical image alignment and show that coarse-to-fine registration is also applicable to NeRF.
Furthermore, we show that na\"ively applying positional encoding in NeRF has a negative impact on registration with a synthesis-based objective.
Experiments on synthetic and real-world data show that BARF can effectively optimize the neural scene representations and resolve large camera pose misalignment at the same time.
This enables view synthesis and localization of video sequences from unknown camera poses, opening up new avenues for visual localization systems (\eg SLAM) and potential applications for dense 3D mapping and reconstruction.

%% file: 1-introduction.tex

\section{Introduction}

Humans have strong capabilities of reasoning about 3D geometry through our vision from the slightest ego-motion.
When watching movies, we can immediately infer the 3D spatial structures of objects and scenes inside the videos.
This is because we have an inherent ability of associating spatial correspondences of the same scene across continuous observations, without having to make sense of the relative camera or ego-motion.
Through pure visual perception, not only can we recover a mental 3D representation of \emph{what} we are looking at, but meanwhile we can also recognize \emph{where} we are looking at the scene from.

\begin{figure}[t!]
    \centering
    \includegraphics[width=1\linewidth,page=1]{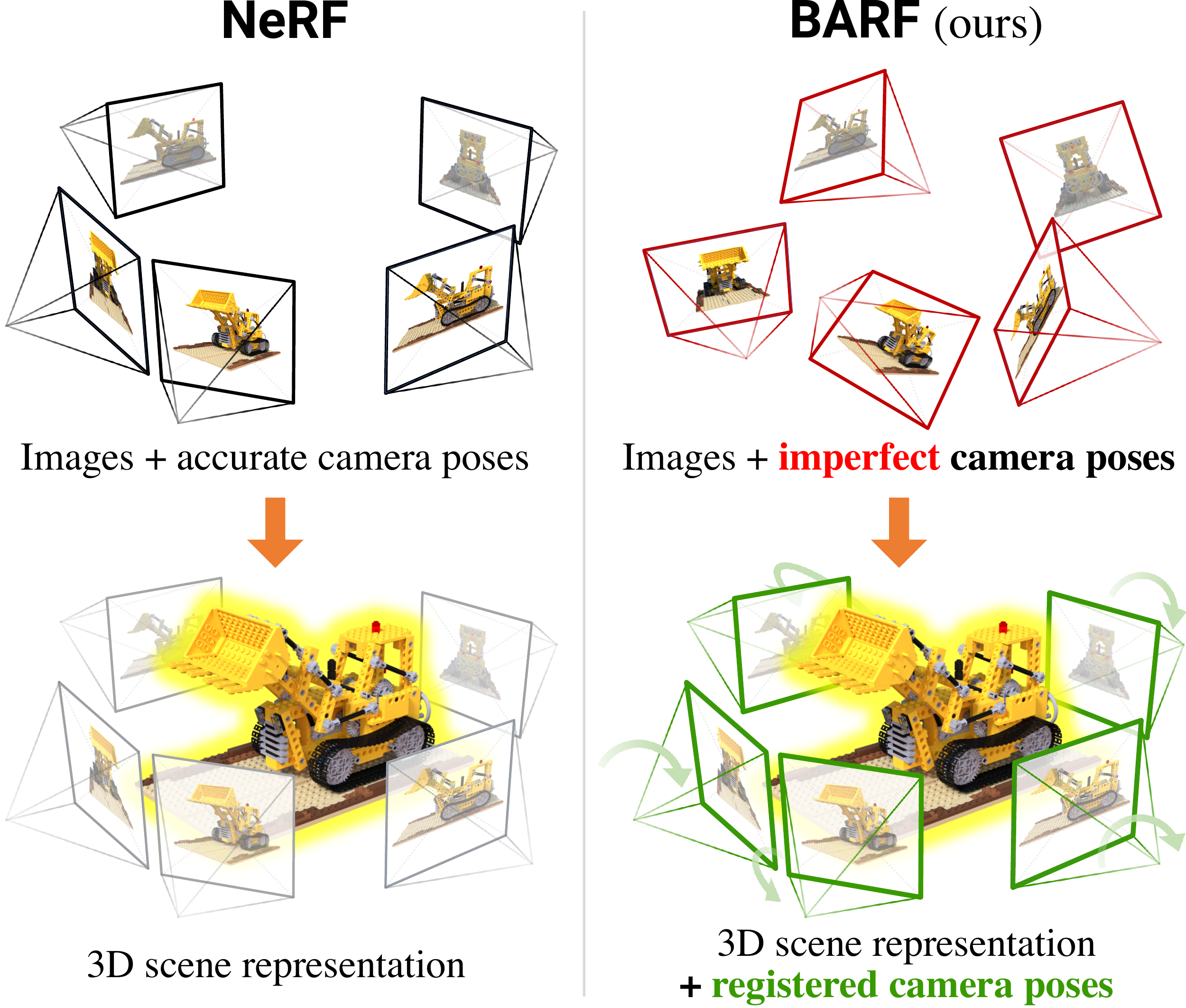}
    \caption{
        Training NeRF requires accurate camera poses for all images.
        We present {\bf BARF} for learning 3D scene representations from \emph{imperfect} (or even \emph{unknown}) camera poses by jointly optimizing for registration and reconstruction.
    }
    \label{fig:teaser}
\end{figure}

Simultaneously solving for the 3D scene representation from RGB images (\ie \textbf{reconstruction}) and localizing the given camera frames (\ie \textbf{registration}) is a long-standing chicken-and-egg problem in computer vision --- recovering the 3D structure requires observations with known camera poses, while localizing the cameras requires reliable correspondences from the reconstruction.
Classical methods such as structure from motion (\SFM)~\cite{hartley2004,schonberger2016structure} or SLAM~\cite{engel2014lsd,mur2015orb} approach this problem through local registration followed by global geometric bundle adjustment (BA) on both the structure and cameras.
\SFM and SLAM systems, however, are sensitive to the quality of local registration and easily fall into suboptimal solutions.
In addition, the sparse nature of output 3D point clouds (often noisy) limits downstream vision tasks that requires dense geometric reasoning.

Closely related to 3D reconstruction from imagery is the problem of view synthesis.
Though not primarily purposed for recovering explicit 3D structures, recent advances on photorealistic view synthesis have opted to recover an intermediate dense 3D-aware representation (\eg depth~\cite{flynn2019deepview,wiles2020synsin}, multi-plane images~\cite{zhou2018stereo,srinivasan2019pushing,tucker2020single}, or volume density~\cite{lombardi2019neural,mildenhall2020nerf}), followed by neural rendering techniques~\cite{eslami2018neural,meshry2019neural,sitzmann2019scene,tewari2020state} to synthesize the target images.
In particular, Neural Radiance Fields (NeRF)~\cite{mildenhall2020nerf} have demonstrated its remarkable ability for high-fidelity view synthesis.
NeRF encodes 3D scenes with a neural network mapping 3D point locations to color and volume density.
This allows the scenes to be represented with compact memory footprint without limiting the resolution of synthesized images.
The optimization process of the network is constrained to obey the principles of classical volume rendering~\cite{levoy1990efficient}, making the learned representation interpretable as a continuous 3D volume density function.

Despite its notable ability for photorealistic view synthesis and 3D scene representation, a hard prerequisite of NeRF (as well as other view synthesis methods) is accurate camera poses of the given images, which is typically obtained through auxiliary off-the-shelf algorithms.
One straightforward way to circumvent this limitation is to additionally optimize the pose parameters with the NeRF model via backpropagation.
As discussed later in the paper, however, na\"ive pose optimization with NeRF is sensitive to initialization. It may lead to suboptimal solutions of the 3D scene representation, degrading the quality of view synthesis.

In this paper, we address the problem of training NeRF representations from imperfect camera poses --- the joint problem of \emph{reconstructing} the 3D scene and \emph{registering} the camera poses (Fig.~\ref{fig:teaser}).
We draw inspiration from the success of classical image alignment methods and establish a theoretical connection, showing that coarse-to-fine registration is also critical to NeRF.
Specifically, we show that positional encoding~\cite{vaswani2017attention} of input 3D points plays a crucial role --- as much as it enables fitting to high-frequency functions~\cite{tancik2020fourier}, positional encoding is also more susceptible to suboptimal registration results.
To this end, we present Bundle-Adjusting NeRF (BARF), a simple yet effective strategy for coarse-to-fine registration on coordinate-based scene representations.
BARF can be regarded as a type of \emph{photometric} BA~\cite{delaunoy2014photometric,alismail2016photometric,lin2019photometric} using view synthesis as the proxy objective.
Unlike traditional BA, however, BARF can learn scene representations \emph{from scratch} (\ie from randomly initialized network weights), lifting the reliance of local registration subprocedures and allowing for more generic applications.

In summary, we present the following contributions:

\vspace{-2pt}
\begin{itemize}[leftmargin=24pt]
	\setlength\itemsep{0pt}
	\item We establish a theoretical connection between classical image alignment to joint registration and reconstruction with Neural Radiance Fields (NeRF).
	\item We show that susceptibility to noise from positional encoding affects the basin of attraction for registration, and we present a simple strategy for coarse-to-fine registration on coordinate-based scene representations.
	\item Our proposed BARF can successfully recover scene representations from imperfect camera poses, allowing for applications such as view synthesis and localization of video sequences from unknown poses.
\end{itemize}

%% file: 2-relatedwork.tex

\section{Related Work}

\noindent\textbf{Structure from motion (\SFM) and SLAM.}
Given a set of input images, \SFM~\cite{pollefeys1999self,pollefeys2004visual,snavely2006photo,snavely2008modeling,agarwal2011building,wu2011visualsfm} and SLAM~\cite{newcombe2011dtam,engel2014lsd,mur2015orb,yang2021asynchronous} systems aim to recover the 3D structure and the sensor poses simultaneously.
These can be classified into (a) \emph{indirect} methods that rely on keypoint detection and matching~\cite{davison2007monoslam,mur2015orb} and (b) \emph{direct} methods that exploit photometric consistency~\cite{alismail2016photometric,engel2017direct}.
Modern pipelines following the indirect route have achieved tremendous success~\cite{schonberger2016structure}; however, they often suffer at textureless regions and repetitive patterns, where distinctive keypoints cannot be reliably detected.
Researchers have thus sought to use neural networks to learn discriminative features directly from data~\cite{detone2018superpoint,ono2018lf,dusmanu2019d2}.

Direct methods, on the other hand, do not rely on such distinctive keypoints --- every pixel can contribute to maximizing photometric consistency, leading to improved robustness in sparsely textured environments~\cite{wang2017stereo}.
They can also be naturally integrated into deep learning frameworks through image reconstruction losses~\cite{zhou2017unsupervised,wang2018learning,yin2018geonet}.
Our method BARF lies under the broad umbrella of direct methods, as BARF learns 3D scene representations from RGB images while also localizing the respective cameras.
However, unlike classical \SFM and SLAM that represent 3D structures with explicit geometry (\eg point clouds), BARF encodes the scenes as coordinate-based representations with neural networks.

\vspace{4pt}
\noindent\textbf{View synthesis.}
Given a set of posed images, view synthesis attempts to simulate how a scene would look like from novel viewpoints~\cite{chen1993view,levoy1996light,szeliski1998stereo,heigl1999plenoptic}.
The task has been closely tied to 3D reconstruction since its introduction~\cite{debevec1996modeling,zitnick2004high,hedman2017casual}.
Researchers have investigated blending pixel colors based on depth maps~\cite{chaurasia2013depth} or leveraging proxy geometry to warp and composite the synthesized image~\cite{kopf2014first}.
However, since the problem is inherently ill-posed, there are still multiple restrictions and assumptions on the synthesized viewpoints.

State-of-the-art methods have capitalized on neural networks to learn both the scene geometry and statistical priors from data.
Various representations have been explored in this direction, \eg depth~\cite{flynn2019deepview,wiles2020synsin,riegler2020free,riegler2020stable}, layered depth~\cite{tulsiani2018layer,shih20203d}, multi-plane images~\cite{zhou2018stereo,srinivasan2019pushing,tucker2020single}, volume density~\cite{lombardi2019neural,mildenhall2020nerf}, and mesh sheets~\cite{hu2020worldsheet}.
Unfortunately, these view synthesis methods still require the camera poses to be known \emph{a priori}, largely limiting their applications in practice.
In contrast, our method BARF is able to effectively learn 3D representations that encodes the underlying scene geometry from imperfect or even unknown camera poses.

\vspace{4pt}
\noindent\textbf{Neural Radiance Fields (NeRF).}
Recently, Mildenhall~\etal~\cite{mildenhall2020nerf} proposed NeRF to synthesize novel views of static, complex scenes from a set of posed input images.
The key idea is to model the continuous radiance field of a scene with a multi-layer perceptron (MLP), followed by differentiable volume rendering to synthesize the images and backpropagate the photometric errors.
NeRF has drawn wide attention across the vision community~\cite{zhang2020nerf++,niemeyer2020giraffe,rebain2020derf,park2020deformable,yen2020inerf} due to its simplicity and extraordinary performance.
It has also been extended on many fronts, \eg reflectance modeling for photorealistic relighting~\cite{boss2020nerd,srinivasan2020nerv} and dynamic scene modeling that integrates the motion of the world~\cite{li2020neural,xian2020space,pumarola2020d}.
Recent works have also sought to exploit a large corpus of data to pretrain the MLP, enabling the ability to infer the radiance field from a single image~\cite{gao2020portrait,yu2020pixelnerf,rematas2021sharf,schwarz2020graf}.

While impressive results have been achieved by the above NeRF-based models, they have a common drawback --- the requirement of \emph{posed} images.
Our proposed BARF allows us to circumvent such requirement.
We show that with a simple coarse-to-fine bundle adjustment technique, we can recover from imperfect camera poses (including \emph{unknown} poses of video sequences) and learn the NeRF representation simultaneously.
Concurrent to our work, NeRF-{}-~\cite{wang2021nerf} introduced an empirical, two-stage pipeline to estimate unknown camera poses.
Our method BARF, in contrast, is motivated by mathematical insights and can recover the camera poses within a single course of optimization, allowing for direct utilities for various NeRF applications and extensions.

%% file: 3-approach.tex

\section{Approach} \label{sec:approach}

We unfold this paper by motivating with the simpler 2D case of classical image alignment as an example.
Then we discuss how the same concept is also applicable to the 3D case, giving inspiration to our proposed BARF.


\subsection{Planar Image Alignment (2D)} \label{sec:image}

Let $\x \in \Real^2$ be the 2D pixel coordinates and $\I: \Real^2 \to \Real^3$ be the imaging function.
Image alignment aims to find the relative geometric transformation which minimizes the photometric error between two images $\I_1$ and $\I_2$.
The problem can be formulated with a synthesis-based objective:
\begin{align} \label{eq:2d}
    \min_{\p} \sum_{\x} \eucnorm{ \I_1(\W(\x;\p)) - \I_2(\x) }^2 \;,
\end{align}
where $\W: \Real^2 \to \Real^2$ is the warp function parametrized by $\p \in \Real^P$ (with $P$ as the dimensionality).
As this is a nonlinear problem,
gradient-based optimization is the method of choice: given the current warp state $\p$, warp updates $\delp$ are iteratively solved for and updated to the solution via $\p \gets \p + \delp$. 
Here, $\delp$ can be written in a generic form of
\begin{align} \label{eq:update}
    \resizebox{0.9\linewidth}{!}{$ \displaystyle
        \!\! \delp = - \A(\x;\p) \sum_{\x}  \J(\x;\p)^{\!\top\!} \big( \I_1(\W(\x;\p)) - \I_2(\x) \big) \;, \!\!
    $}
\end{align}
where $\J \in \Real^{3\times P}$ is termed the steepest descent image, and $\A$ is a generic transformation which depends on the choice of the optimization algorithm.
The seminal Lucas-Kanade algorithm~\cite{lucas1981iterative} approaches the problem using Gauss-Newton optimization, \ie $\A(\x;\p) = (\sum_{\x} \J(\x;\p)^\top\J(\x;\p))^{-1}$; alternatively, one could also choose first-order optimizers such as (stochastic) gradient descent which can be more naturally incorporated into modern deep learning frameworks, where $\A$ would correspond to a scalar learning rate.

\begin{figure}[t!]  
    \centering
    \includegraphics[width=1\linewidth,page=1]{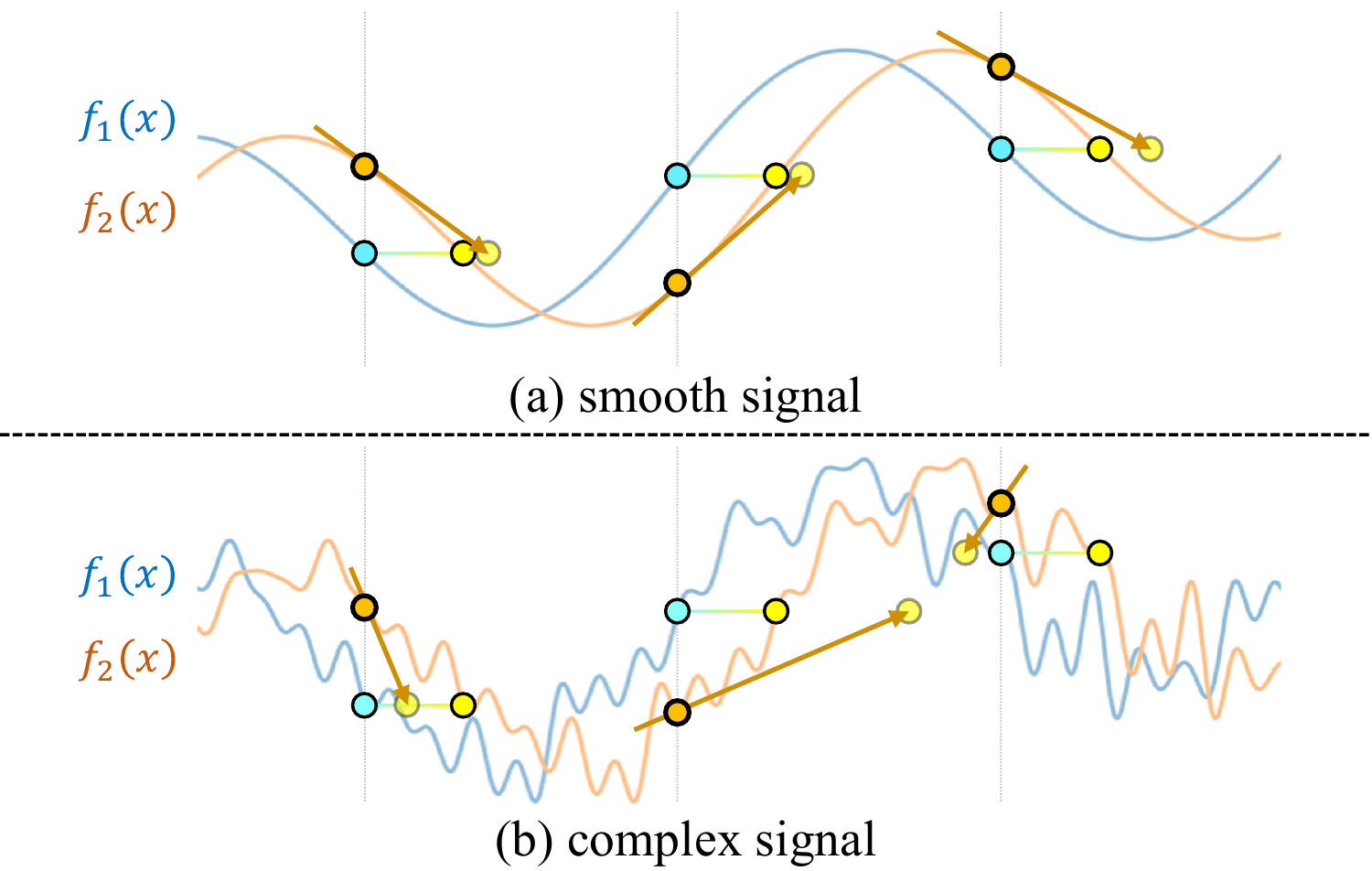}
    \caption{
        {\bf Predicting alignment from signal differences.}
        Consider two 1D signals where $f_1(x) = f_2(x+c)$ differs by an offset $c$.
        When solving for alignment, smoother signals can predict more coherent displacements than complex signals, which easily results in suboptimal alignment.
    }
    \label{fig:interpolate}
\end{figure}

The steepest descent image $\J$ can be expanded as
\begin{align} \label{eq:jacobian}
    \J(\x;\p) = \pderiv{\I_1(\W(\x;\p))}{\W(\x;\p)} \pderiv{\W(\x;\p)}{\p} \;,
\end{align}
where $\pderiv{\W(\x;\p)}{\p} \in \Real^{2\times P}$ is the warp Jacobian constraining the pixel displacements with respect to the predefined warp.
At the heart of gradient-based registration are the image gradients $\pderiv{\I(\x)}{\x} \in \Real^{3\times 2}$ modeling a local per-pixel linear relationship between appearance and spatial displacements, which is classically estimated via finite differencing.
The overall warp update $\delp$ can be more effectively estimated from pixel value differences if the per-pixel predictions are coherent (Fig.~\ref{fig:interpolate}), \ie the image signals are smooth.
However, as natural images are typically complex signals,
gradient-based registration on raw images is susceptible to suboptimal solutions if poorly initialized.
Therefore, coarse-to-fine strategies have been practiced by blurring the images at earlier stages of registration, effectively widening the basin of attraction and smoothening the alignment landscape.

\vspace{6px}
\noindent\textbf{Images as neural networks.}
An alternative formulation of the problem is to learn a coordinate-based image representation with a neural network while also solving for the warp $\p$.
Writing the network as $f: \Real^2 \to \Real^3$ and denoting $\bTheta$ as its parameters, one can instead choose to optimize the objective
\begin{align} \label{eq:2d_b}
    \min_{\p,\bTheta} \sum_{\x} \big( &\eucnorm{ f(\x;\bTheta) - \I_1(\x) }^2 \nonumber \\[-6pt]
                                    + &\eucnorm{ f(\W(\x;\p);\bTheta) - \I_2(\x) }^2 \big) \;,
\end{align}
or alternatively, one may choose to solve for warp parameters $\p_1$ and $\p_2$ respectively for both images $\I_1$ and $\I_2$ through
\begin{align} \label{eq:2d_c}
    \min_{\p_1,\p_2,\bTheta} \; \sum_{i=1}^{M} \sum_{\x} &\eucnorm{ f(\W(\x;\p_i);\bTheta) - \I_i(\x) }^2 \;,
\end{align}
where $M=2$ is the number of images.
Albeit similar to~\eqref{eq:2d}, the image gradients become the analytical Jacobian of the network $\pderiv{f(\x)}{\x}$ instead of numerical estimation.
By manipulating the network $f$, this also enables more principled control of the signal smoothness for alignment without having to rely on heuristic blurring on images, making these forms generalizable to 3D scene representations (Sec.~\ref{sec:nerf}).


\setlength{\abovedisplayskip}{7.5pt}
\setlength{\belowdisplayskip}{7.5pt}

\subsection{Neural Radiance Fields (3D)} \label{sec:nerf}

We discuss the 3D case of recovering the 3D scene representation from Neural Radiance Fields (NeRF)~\cite{mildenhall2020nerf} \emph{jointly} with the camera poses.
To signify the analogy to Sec.~\ref{sec:image}, we deliberately overload the notations $\x$ as 3D points, $\W$ as camera pose transformations, and $f$ as the network in NeRF.

NeRF encodes a 3D scene as a continuous 3D representation using an MLP $f: \Real^3 \to \Real^4$ to predict the RGB color $\c \in \Real^3$ and volume density $\sigma \in \Real$ for each input 3D point $\x \in \Real^3$.
This can be summarized as $\y = [\c;\sigma]^\top = f(\x;\bTheta)$, where $\bTheta$ is the network parameters\footnotemark.
NeRF assumes an emission-only model, \ie the rendered color of a pixel is dependent only on the emitted radiance of 3D points along the viewing ray, without considering external lighting factors.

We first formulate the rendering operation of NeRF in the camera view space.
Given pixel coordinates $\u \in \Real^2$ and denoting its homogeneous coordinates as $\uhom = [\u;1]^\top \in \Real^3$, we can express a 3D point $\x_i$ along the viewing ray at depth $z_i$ as $\x_i = z_i \uhom$.
The RGB color $\Ihat$ at pixel location $\u$ is extracted by volume rendering via
\footnotetext{In practice, $f$ is also conditioned on the viewing direction~\cite{mildenhall2020nerf} for modeling view-dependent effects, which we omit here for simplicity.}
\begin{align} \label{eq:volrender}
    \Ihat(\u) = \int_{z_\text{near}}^{z_\text{far}} T(\u,z) \sigma(z \uhom) \c(z \uhom) \mathrm{d}z \;,
\end{align}
where $T(\u,z) = \exp\!\big(\!-\!\int_{z_\text{near}}^{z} \sigma(z' \uhom) \mathrm{d}z' \big)$, and $z_\text{near}$ and $z_\text{far}$ are bounds on the depth range of interest.
We refer our readers to Levoy~\cite{levoy1990efficient} and Mildenhall~\etal~\cite{mildenhall2020nerf} for a more detailed treatment on volume rendering.
In practice, the above integral formulations are approximated numerically via quadrature on discrete $N$ points at depth $\{z_1,\dots,z_N\}$ sampled along the ray.
This involves $N$ evaluations of the network $f$, whose output $\{\y_1,\dots,\y_N\}$ are further composited through volume rendering.
We can summarize the ray compositing function as $g: \Real^{4N} \to \Real^3$ and rewrite $\Ihat(\u)$ as $\Ihat(\u) = g\left(\y_1,\dots,\y_N\right)$.
Note that $g$ is differentiable but deterministic, \ie there are no learnable parameters associated.

Under a 6-DoF camera pose parametrized by $\p \in \Real^6$, a 3D point $\x$ in the camera view space can be transformed to the 3D world coordinates through a 3D rigid transformation $\W: \Real^3 \to \Real^3$.
Therefore, the synthesized RGB value at pixel $\u$ becomes a function of the camera pose $\p$ as
\begin{align} \label{eq:comp2}
    \resizebox{0.9\linewidth}{!}{$ \displaystyle
        \!\! \Ihat(\u;\p) = g\Big(f(\W(z_1 \uhom;\p);\bTheta),\dots,f(\W(z_N \uhom;\p);\bTheta)\Big) . \!\!
    $}
\end{align}
Given $M$ images $\{\I_i\}_{i=1}^M$, our goal is to optimize NeRF \emph{and} the camera poses $\{\p_i\}_{i=1}^M$ over the synthesis-based objective
\begin{align} \label{eq:loss}
    \min_{\p_1, \dots, \p_M, \bTheta} \; \sum_{i=1}^{M} \sum_{\u} \big\| \Ihat(\u; \p_i, \bTheta) - \I_i(\u) \big\|_2^2 \;,
\end{align}
where $\Ihat$ also depends on the network parameters $\bTheta$.

One may notice the analogy between the synthesis-based objectives of 2D image alignment~\eqref{eq:2d_c} and NeRF~\eqref{eq:loss}.
Similarly, we can also derive the ``steepest descent image'' as
\begin{align} \label{eq:taylor3d_b}
    \resizebox{0.88\linewidth}{!}{$ \displaystyle
        \!\! \J(\u;\p) = \sum_{i=1}^{N} \pderiv{g(\y_1,\dots,\y_N)}{\y_i} \pderiv{\y_i(\p)}{\x_i(\p)} \pderiv{\W(z_i \uhom;\p)}{\p} \;, \!\!
    $}
\end{align}
which is formed via backpropagation in practice.
The linearization~\eqref{eq:taylor3d_b} is also analogous to the 2D case of~\eqref{eq:jacobian}, where the Jacobian of the network $\pderiv{\y}{\x} = \pderiv{f(\x)}{\x}$ linearly relates the change of color $\c$ and volume density $\sigma$ with 3D spatial displacements.
To solve for effective camera pose updates $\delp$ through backpropagation, it is also desirable to control the smoothness of $f$ for predicting coherent geometric displacements from the sampled 3D points $\{\x_1,\dots,\x_N\}$.


\subsection{On Positional Encoding and Registration} \label{sec:posenc}

The key of enabling NeRF to synthesize views with high fidelity is positional encoding~\cite{vaswani2017attention}, a deterministic mapping of input 3D coordinates $\x$ to higher dimensions of different sinusoidal frequency bases\footnotemark.
We denote $\gamma: \Real^3 \to \Real^{3+6L}$ as the positional encoding with $L$ frequency bases, defined as
\begin{align} \label{eq:posenc}
    \gamma(\x) = \big[ \x, \gamma_0(\x), \gamma_1(\x), \dots, \gamma_{L-1}(\x) \big] \in \Real^{3+6L}\;,
\end{align}
where the $k$-th frequency encoding $\gamma_k(\x)$ is
\begin{align} \label{eq:posenc-1}
    \gamma_k(\x) = \big[ \cos(2^{k}\pi\x), \sin(2^{k}\pi\x) \big] \in \Real^{6} \;,
\end{align}
with the sinusoidal functions operating coordinate-wise.
The special case of $L=0$ makes $\gamma$ an identity mapping function.
The network $f$ is thus a composition of $f(\x) = f' \circ \gamma(\x)$, where $f'$ is the subsequent learnable MLP.
Positional encoding allows coordinate-based neural networks, which are typically bandwidth limited, to represent signals of higher frequency with faster convergence behaviors~\cite{tancik2020fourier}.
\footnotetext{Although we focus on 3D input coordinates here, positional encoding is also directly applicable to 2D image coordinates in Sec.~\ref{sec:image} as well.}

The Jacobian of the $k$-th positional encoding $\gamma_k$ is
\begin{align} \label{eq:posenc-jac}
    \pderiv{\gamma_k(\x)}{\x} = 2^{k}\pi \cdot \big[ -\sin(2^{k}\pi\x), \cos(2^{k}\pi\x) \big] \;,
\end{align}
which amplifies the gradient signals from the MLP $f'$ by $2^{k}\pi$ with its direction changing at the same frequency.
This makes it difficult to predict effective updates $\delp$, since gradient signals from the sampled 3D points are incoherent (in terms of both direction and magnitude) and can easily cancel out each other.
Therefore, na\"ively applying positional encoding can become a double-edged sword to NeRF for the task of joint registration and reconstruction.


\subsection{Bundle-Adjusting Neural Radiance Fields} \label{sec:barf}

We describe our proposed BARF, a simple yet effective strategy for coarse-to-fine registration for NeRF.
The key idea is to apply a smooth mask on the encoding at different frequency bands (from low to high) over the course of optimization, which acts like a dynamic low-pass filter.
Inspired by recent work of learning coarse-to-fine deformation flow fields~\cite{park2020deformable}, we weigh the $k$-th frequency component of $\gamma$ as
\begin{align} \label{eq:weight}
    \gamma_k(\x;\alpha) = w_k(\alpha) \cdot \big[ \cos(2^{k}\pi\x), \sin(2^{k}\pi\x) \big] \;,
\end{align}
where the weight $w_k$ is defined as
\begin{align} \label{eq:window}
    \resizebox{0.8\linewidth}{!}{$ \displaystyle
        w_k(\alpha) = 
        \begin{cases}
            0 & \text{if}~~ \alpha < k \vspace{-1pt} \\
            \displaystyle \frac{1-\cos((\alpha-k)\pi)}{2} & \text{if}~~ 0 \leq \alpha-k < 1 \vspace{-1pt} \\
            1 & \text{if}~~ \alpha-k \geq 1
        \end{cases}
    $}
\end{align}
and $\alpha \in [0,L]$ is a controllable parameter proportional to the optimization progress.
The Jacobian of $\gamma_k$ thus becomes
\begin{align} \label{eq:posenc-jac-window}
    \resizebox{0.88\linewidth}{!}{$ \displaystyle
        \!\! \pderiv{\gamma_k(\x;\alpha)}{\x} = w_k(\alpha) \cdot 2^{k}\pi \cdot \big[ -\sin(2^{k}\pi\x), \cos(2^{k}\pi\x) \big] \;. \!\!
    $}
\end{align}
When $w_k(\alpha)=0$, the contribution to the gradient from the $k$-th (and higher) frequency component is nullified.

Starting from the raw 3D input $\x$ ($\alpha=0$), we gradually activate the encodings of higher frequency bands until full positional encoding is enabled ($\alpha=L$), equivalent to the original NeRF model.
This allows BARF to discover the correct registration with an initially smooth signal and later shift focus to learning a high-fidelity scene representation.

%% file: 4-experiments.tex

\newcommand{\?}{\filler\bf ???}

\section{Experiments}

We validate the effectiveness of our proposed BARF with a simple experiment of 2D planar image alignment, and show how the same coarse-to-fine registration strategy can be generalized to NeRF~\cite{mildenhall2020nerf} for learning 3D scene representations.


\subsection{Planar Image Alignment (2D)} \label{sec:exp-image}

We choose a representative image from ImageNet~\cite{deng2009imagenet}, shown in Fig.~\ref{fig:planar}.
Given $M=5$ patches from the image generated with homography perturbations (Fig.~\ref{fig:planar}(a)), we aim to find the homography warp parameters $\p \in \Real^8$ for each patch (Fig.~\ref{fig:planar}(b)) while \emph{also} learning the neural representation of the entire image with a network $f$ by optimizing~\eqref{eq:2d_c}.
We initialize all $M$ patches with a center crop (Fig.~\ref{fig:planar}(c)), and we anchor the warp of the first patch as identity so the recovered image would be implicitly aligned to the raw image.
We parametrize homography warps with the $\mathfrak{sl}(3)$ Lie algebra.

\vspace{4pt}
\noindent\textbf{Experimental settings.}
We investigate how positional encoding impacts this problem by comparing networks with na\"ive (full) positional encoding and without any encoding.
We use a simple ReLU MLP for $f$ with four 256-dimensional hidden units, and we use the Adam optimizer~\cite{kingma2014adam} to optimize both the network weights and the warp parameters for $5000$ iterations with a learning rate of $0.001$.
For BARF, we linearly adjust $\alpha$ for the first $2000$ iterations and activate all frequency bands ($L=8$) for the remaining iterations.

\vspace{4pt}
\noindent\textbf{Results.}
We visualize the registration results in Fig.~\ref{fig:planar-results}.
Alignment with full positional encoding results in suboptimal registration with ghostly artifacts in the recovered image representation.
On the other hand, alignment without positional encoding achieves decent registration results, but cannot recover the image with sufficient fidelity.
BARF discovers the precise geometric warps with the image representation optimized with high fidelity,
quantitatively reflected in Table~\ref{table:planar}.
The image alignment experiment demonstrates the general advantage of BARF for coordinate-based representations.

\begin{table*}[t!]
    \centering
    \begin{minipage}{0.415\linewidth}
        \includegraphics[width=1\linewidth,page=1]{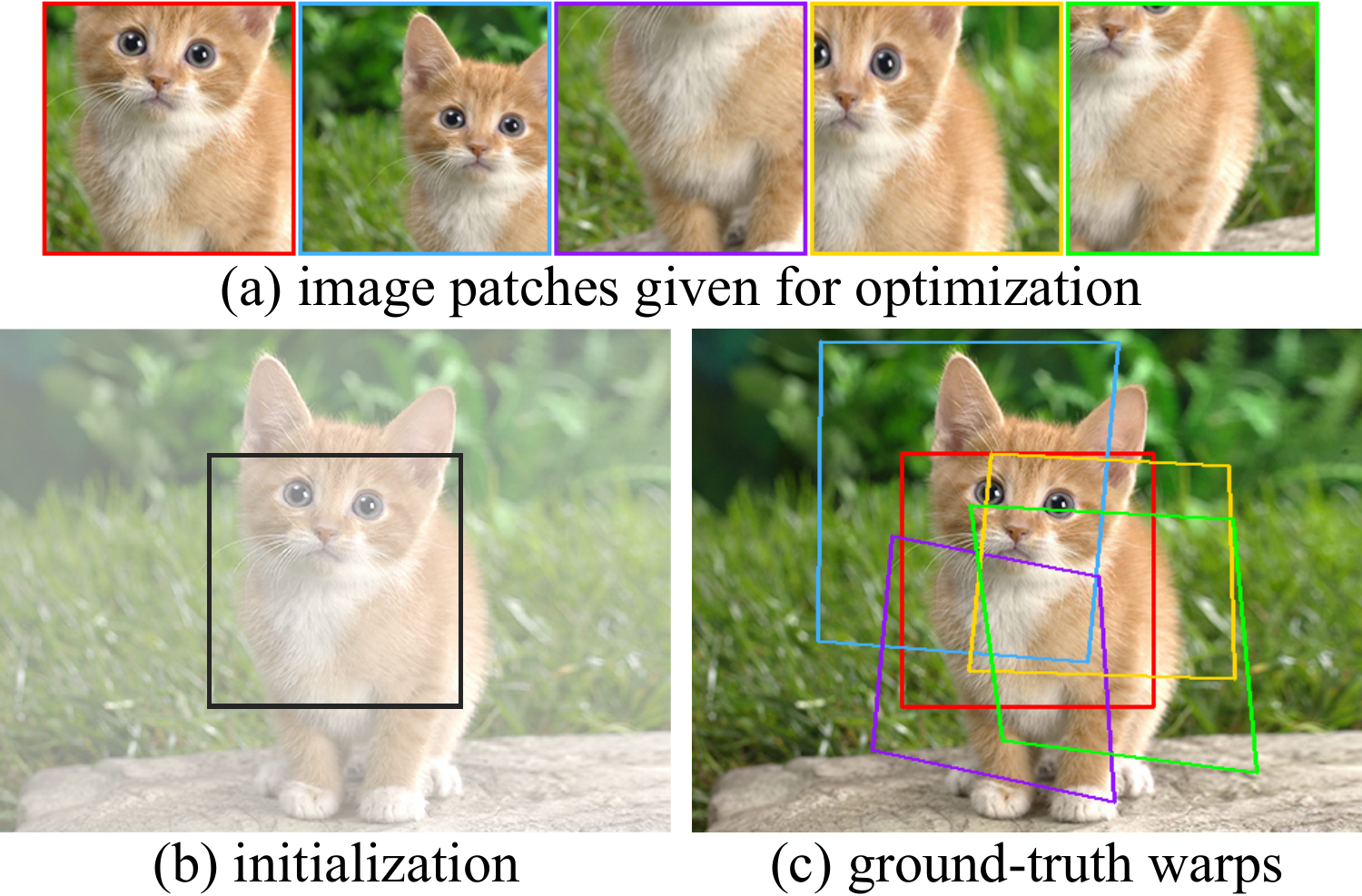}
        \vspace{-16pt}
        \captionof{figure}{
            Given image patches color-coded in (a), we aim to recover the alignment \emph{and} the neural representation of the entire image, with the patches initialized to center crops shown in (b) and the corresponding ground-truth warps shown in (c).
        }
        \label{fig:planar}
        \vspace{12pt}
        \centering
        \resizebox{0.9\linewidth}{!}{
            \begin{tabular}{c|c|c}
                \toprule
                positional encoding & $\mathfrak{sl}(3)$ error & patch PSNR \\
                \midrule
                na\"ive (full) & 0.2949 & 23.41 \\
                without & 0.0641 & 24.72 \\
                BARF (coarse-to-fine) & 0.0096 & 35.30 \\
                \bottomrule 
            \end{tabular}
        }
        \vspace{-4pt}
        \captionof{table}{
            {\bf Quantitative results} of planar image alignment.
            BARF optimizes for more accurate alignment and patch reconstruction compared to the baselines.
        }
        \label{table:planar}
    \end{minipage}
    \hspace{8pt}
    \begin{minipage}{0.555\linewidth}
        \includegraphics[width=1\linewidth,page=1]{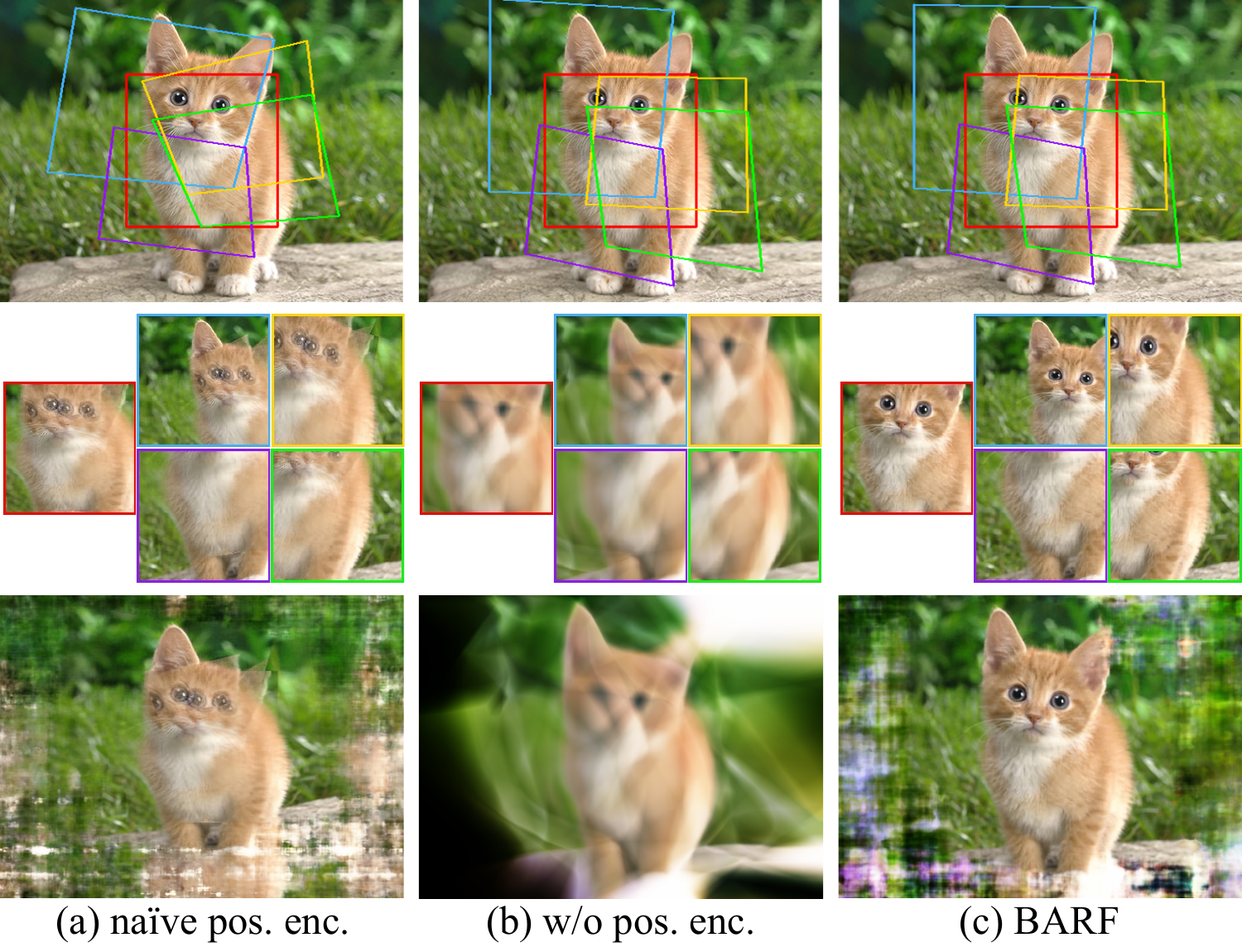}
        \captionof{figure}{
            {\bf Qualitative results} of the planar image alignment experiment.
            We visualize the optimized warps (top row), the patch reconstructions in corresponding colors (middle row), and recovered image representation from $f$ (bottom row).
            BARF is able to recover accurate alignment and high-fidelity image reconstruction, while baselines result in suboptimal alignment with na\"ive positional encoding and blurry reconstruction without any encoding.
            Best viewed in color.
        }
        \label{fig:planar-results}
    \end{minipage}
\end{table*}


\subsection{NeRF (3D): Synthetic Objects} \label{sec:exp-nerf-blender}

We investigate the problem of learning 3D scene representations with Neural Radiance Fields (NeRF)~\cite{mildenhall2020nerf} from imperfect camera poses.
We experiment with the 8 synthetic object-centric scenes provided by Mildenhall~\etal~\cite{mildenhall2020nerf}, which consists of $M=100$ rendered images with ground-truth camera poses for each scene for training.

\begin{figure}[t!]
    \centering  
    \includegraphics[width=1\linewidth,page=1]{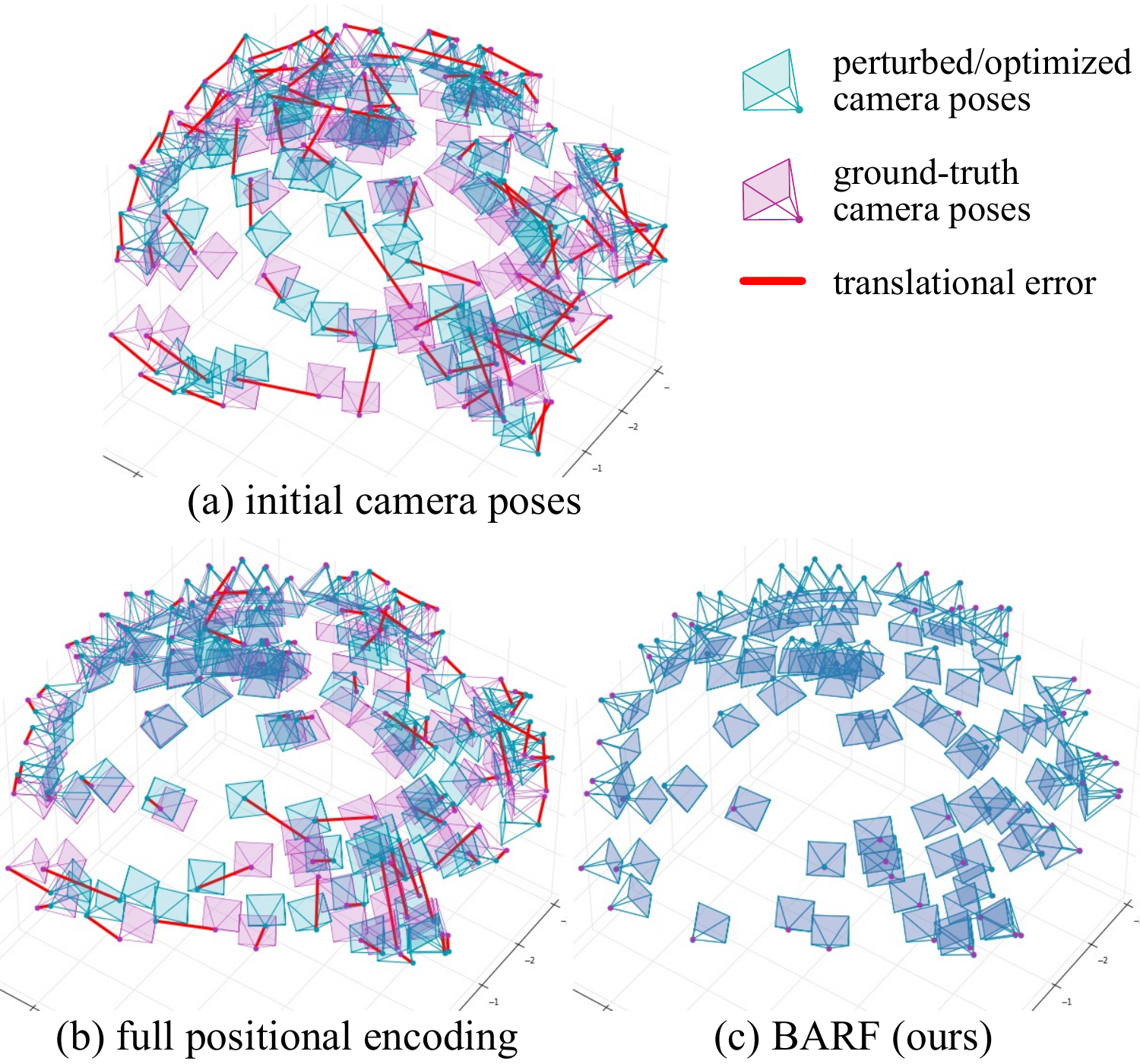}
    \caption{
        Visual comparison of the initial and optimized camera poses (Procrustes aligned) for the \textit{chair} scene.
        BARF successfully realigns all the camera frames while NeRF na\"ive positional encoding gets stuck at suboptimal solutions.
    }
    \label{fig:nerf-camera-blender}
    \vspace{-8pt}
\end{figure}

\vspace{4pt}
\noindent\textbf{Experimental settings.}
We parametrize the camera poses $\p$ with the $\mathfrak{se}(3)$ Lie algebra and assume known intrinsics.
For each scene, we synthetically perturb the camera poses with additive noise $\delta\p \sim \mathcal{N}(\0,0.15\eye)$, which corresponds to a standard deviation of $14.9\degree$ in rotation and $0.26$ in translational magnitude (Fig.~\ref{fig:nerf-camera-blender}(a)).
We optimize the objective in~\eqref{eq:loss} jointly for the scene representation and the camera poses.
We evaluate BARF mainly against the original NeRF model with na\"ive (full) positional encoding; for completeness, we also compare with the same model without positional encoding.

\begin{figure*}[t!]
    \centering  
    \includegraphics[width=1\linewidth,page=1]{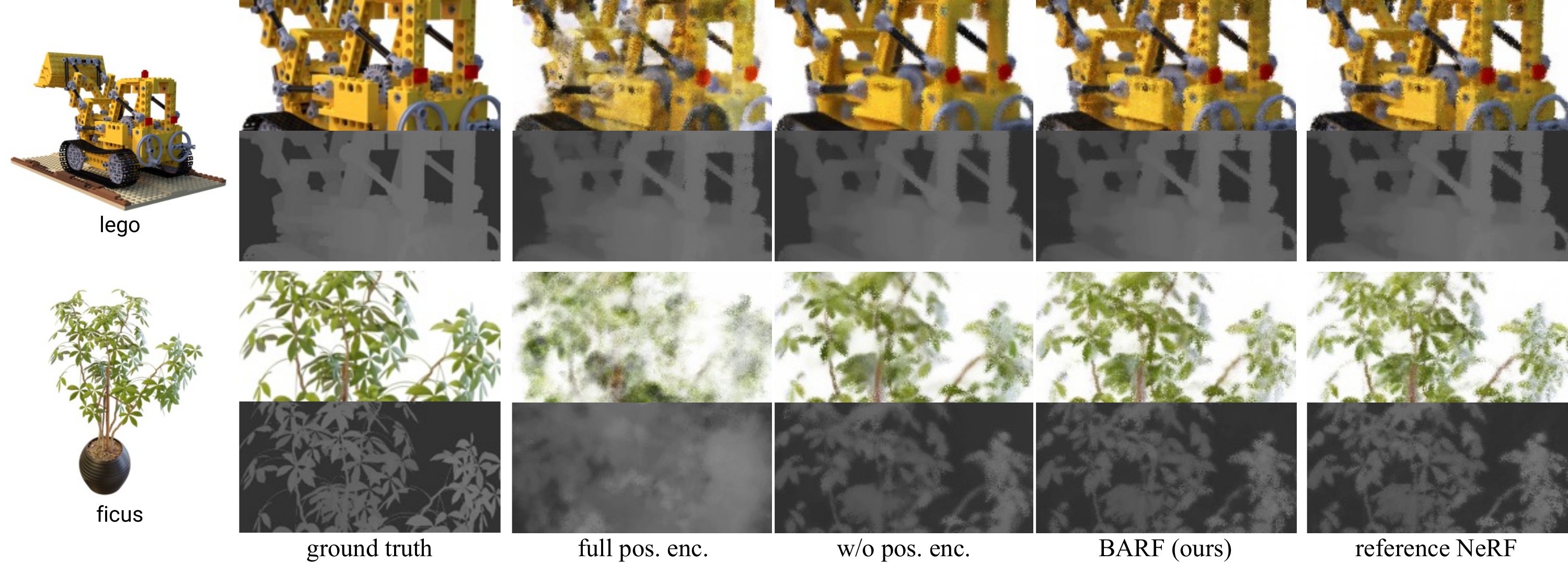}
    \vspace{-20pt}
    \caption{
        {\bf Qualitative results} of NeRF on synthetic scenes.
        We visualize the image synthesis (top) and the expected depth through ray compositing (bottom).
        BARF achieves comparable synthesis quality to the reference NeRF (trained under perfect camera poses), while full positional encoding results in suboptimal registration, leading to synthesis artifacts.
    }
    \label{fig:nerf-results-blender}
\end{figure*}

\vspace{4pt}
\noindent\textbf{Implementation details.}
We follow the architectural settings from the original NeRF~\cite{mildenhall2020nerf} with some modifications.
We train a single MLP with $128$ hidden units in each layer and without additional hierarchical sampling for simplicity.
We resize the images to $400\times 400$ pixels and randomly sample $1024$ pixel rays at each optimization step.
We choose $N=128$ sample for numerical integration along each ray, and we use the softplus activation on the volume density output $\sigma$ for improved stability.
We use the Adam optimizer and train all models for $200$K iterations, with a learning rate of $5\!\times\!10^{-4}$ exponentially decaying to $1\!\times\!10^{-4}$ for the network $f$ and $1\!\times\!10^{-3}$ decaying to $1\!\times\!10^{-5}$ for the poses $\p$.
For BARF, we linearly adjust $\alpha$ from iteration $20$K to $100$K and activate all frequency bands (up to $L=10$) subsequently.

\vspace{4pt}
\noindent\textbf{Evaluation criteria.}
We measure the performance in two aspects: pose error for registration and view synthesis quality for the scene representation.
Since both the scene and camera poses are variable up to a 3D similarity transformation, we evaluate the quality of registration by pre-aligning the optimized poses to the ground truth with Procrustes analysis on the camera locations.
For evaluating view synthesis, we run an additional step of test-time photometric optimization on the trained models~\cite{lin2019photometric,yen2020inerf} to factor out the pose error that may contaminate the view synthesis quality.
We report the average rotation and translation errors for pose and PSNR, SSIM and LPIPS~\cite{zhang2018unreasonable} for view synthesis.

\begin{table*}[t!]
    \centering
    \setlength\tabcolsep{3pt}
    \resizebox{\linewidth}{!}{
        \begin{tabular}{c||ccc|ccc||ccc|c|ccc|c|ccc|c}
            \toprule
            \multirow{4}{*}{Scene} & \multicolumn{6}{c||}{Camera pose registration} & \multicolumn{12}{c}{View synthesis quality} \vspace{1.5pt} \\
            & \multicolumn{3}{c|}{Rotation ($\degree$) $\downarrow$} & \multicolumn{3}{c||}{Translation $\downarrow$} & \multicolumn{4}{c|}{PSNR $\uparrow$} & \multicolumn{4}{c|}{SSIM $\uparrow$} & \multicolumn{4}{c}{LPIPS $\downarrow$} \\
            \cmidrule{2-19}
            & \small full & \small w/o & \small \multirow{2}{*}{BARF}
            & \small full & \small w/o & \small \multirow{2}{*}{BARF}
            & \small full & \small w/o & \small \multirow{2}{*}{BARF} & \small ref.
            & \small full & \small w/o & \small \multirow{2}{*}{BARF} & \small ref.
            & \small full & \small w/o & \small \multirow{2}{*}{BARF} & \small ref. \vspace{-2.5pt} \\
            & \small pos.enc. & \small pos.enc. & \small
            & \small pos.enc. & \small pos.enc. & \small
            & \small pos.enc. & \small pos.enc. & \small & \small NeRF
            & \small pos.enc. & \small pos.enc. & \small & \small NeRF
            & \small pos.enc. & \small pos.enc. & \small & \small NeRF \\
            \midrule
            Chair      &  7.186 & 0.110 & \bf 0.096 & 16.638 & 0.555 & \bf 0.428 & 19.02 & 30.22 & \bf 31.16 & 31.91 & 0.804 & 0.942 & \bf 0.954 & 0.961 & 0.223 & 0.065 & \bf 0.044 & 0.036 \\
            Drums      &  3.208 & 0.057 & \bf 0.043 &  7.322 & 0.255 & \bf 0.225 & 20.83 & 23.56 & \bf 23.91 & 23.96 & 0.840 & 0.893 & \bf 0.900 & 0.902 & 0.166 & 0.116 & \bf 0.099 & 0.095 \\
            Ficus      &  9.368 & 0.095 & \bf 0.085 & 10.135 & \bf 0.430 & 0.474 & 19.75 & 25.58 & \bf 26.26 & 26.68 & 0.836 & 0.922 & \bf 0.934 & 0.941 & 0.182 & 0.070 & \bf 0.058 & 0.051 \\
            Hotdog     &  3.290 & \bf 0.225 & 0.248 &  6.344 & \bf 1.122 & 1.308 & 28.15 & 34.00 & \bf 34.54 & 34.91 & 0.923 & 0.967 & \bf 0.970 & 0.973 & 0.083 & 0.040 & \bf 0.032 & 0.029 \\
            Lego       &  3.252 & 0.108 & \bf 0.082 &  4.841 & 0.391 & \bf 0.291 & 24.23 & 26.35 & \bf 28.33 & 29.28 & 0.876 & 0.880 & \bf 0.927 & 0.942 & 0.102 & 0.112 & \bf 0.050 & 0.037 \\
            Materials  &  6.971 & 0.845 & \bf 0.844 & 15.188 & \bf 2.678 & 2.692 & 16.51 & 26.86 & \bf 27.84 & 28.48 & 0.747 & 0.926 & \bf 0.936 & 0.944 & 0.294 & 0.068 & \bf 0.058 & 0.049 \\
            Mic        & 10.554 & 0.081 & \bf 0.071 & 22.724 & 0.356 & \bf 0.301 & 15.10 & 30.93 & \bf 31.18 & 31.98 & 0.788 & 0.968 & \bf 0.969 & 0.971 & 0.334 & 0.050 & \bf 0.048 & 0.044 \\
            Ship       &  5.506 & 0.095 & \bf 0.075 &  7.232 & 0.354 & \bf 0.326 & 22.12 & 26.78 & \bf 27.50 & 28.00 & 0.755 & 0.833 & \bf 0.849 & 0.858 & 0.255 & 0.175 & \bf 0.132 & 0.118 \\
            \midrule
            Mean       &  6.167 & 0.202 & \bf 0.193 & 11.303 & 0.768 & \bf 0.756 & 22.12 & 26.78 & \bf 27.50 & 29.40 & 0.821 & 0.917 & \bf 0.930 & 0.936 & 0.205 & 0.087 & \bf 0.065 & 0.057 \\
            \bottomrule 
        \end{tabular}
    }
    \vspace{-4pt}
    \caption{
        {\bf Quantitative results} of NeRF on synthetic scenes.
        BARF successfully optimizes for camera registration (with less than $0.2\degree$ rotation error) while still consistently achieving high-quality view synthesis that is comparable to the reference NeRF models (trained under perfect camera poses).
        Translation errors are scaled by $100$.
    }
    \vspace{-2pt}
    \label{table:nerf-blender}
\end{table*}

\vspace{4pt}
\noindent\textbf{Results.}
We visualize the results in Fig.~\ref{fig:nerf-results-blender} and report the quantitative results in Table~\ref{table:nerf-blender}.
BARF takes the best of both worlds of recovering the neural scene representation with the camera pose successfully registered, while na\"ive NeRF with full positional encoding finds suboptimal solutions.
Fig.~\ref{fig:nerf-camera-blender} shows that BARF can achieve near-perfect registration for the synthetic scenes.
Although the NeRF model without positional encoding can also successfully recover alignment, the learned scene representations (and thus the synthesized images) lack the reconstruction fidelity.
As a reference, we also compare the view synthesis quality against standard NeRF models trained under ground-truth poses, showing that BARF can achieve comparable view synthesis quality in all metrics, albeit initialized from imperfect camera poses.


\subsection{NeRF (3D): Real-World Scenes} \label{sec:exp-nerf-llff}

We investigate the challenging problem of learning neural 3D representations with NeRF on real-world scenes, where the camera poses are \emph{unknown}.
We consider the LLFF dataset~\cite{mildenhall2019local}, which consists of 8 forward-facing scenes with RGB images sequentially captured by hand-held cameras.

\vspace{4pt}
\noindent\textbf{Experimental settings.}
We parametrize the camera poses $\p$ with $\mathfrak{se}(3)$ following Sec.~\ref{sec:exp-nerf-blender} but initialize all cameras with the \emph{identity} transformation, \ie $\p_i=\0 \;\; \forall i$.
We assume known camera intrinsics (provided by the dataset).
We compare against the original NeRF model with na\"ive positional encoding, and we use the same evaluation criteria described in Sec.~\ref{sec:exp-nerf-blender}.
However, we note that the camera poses provided in LLFF are also estimations from \SFM packages~\cite{schonberger2016structure}; therefore, the pose evaluation is at most an indication of how well BARF agrees with classical geometric pose estimation.

\vspace{4pt}
\noindent\textbf{Implementation details.}
We follow the same architectural settings from the original NeRF~\cite{mildenhall2020nerf} and resize the images to $480\times 640$ pixels.
We train all models for $200$K iterations and randomly sample $2048$ pixel rays at each optimization step, with a learning rate of $1\!\times\!10^{-3}$ for the network $f$ decaying to $1\!\times\!10^{-4}$, and $3\!\times\!10^{-3}$ for the pose $\p$ decaying to $1\!\times\!10^{-5}$.
We linearly adjust $\alpha$ for BARF from iteration $20$K to $100$K and activate all bands (up to $L=10$) subsequently.

\begin{figure*}[t!]
    \centering  
    \includegraphics[width=0.98\linewidth,page=1]{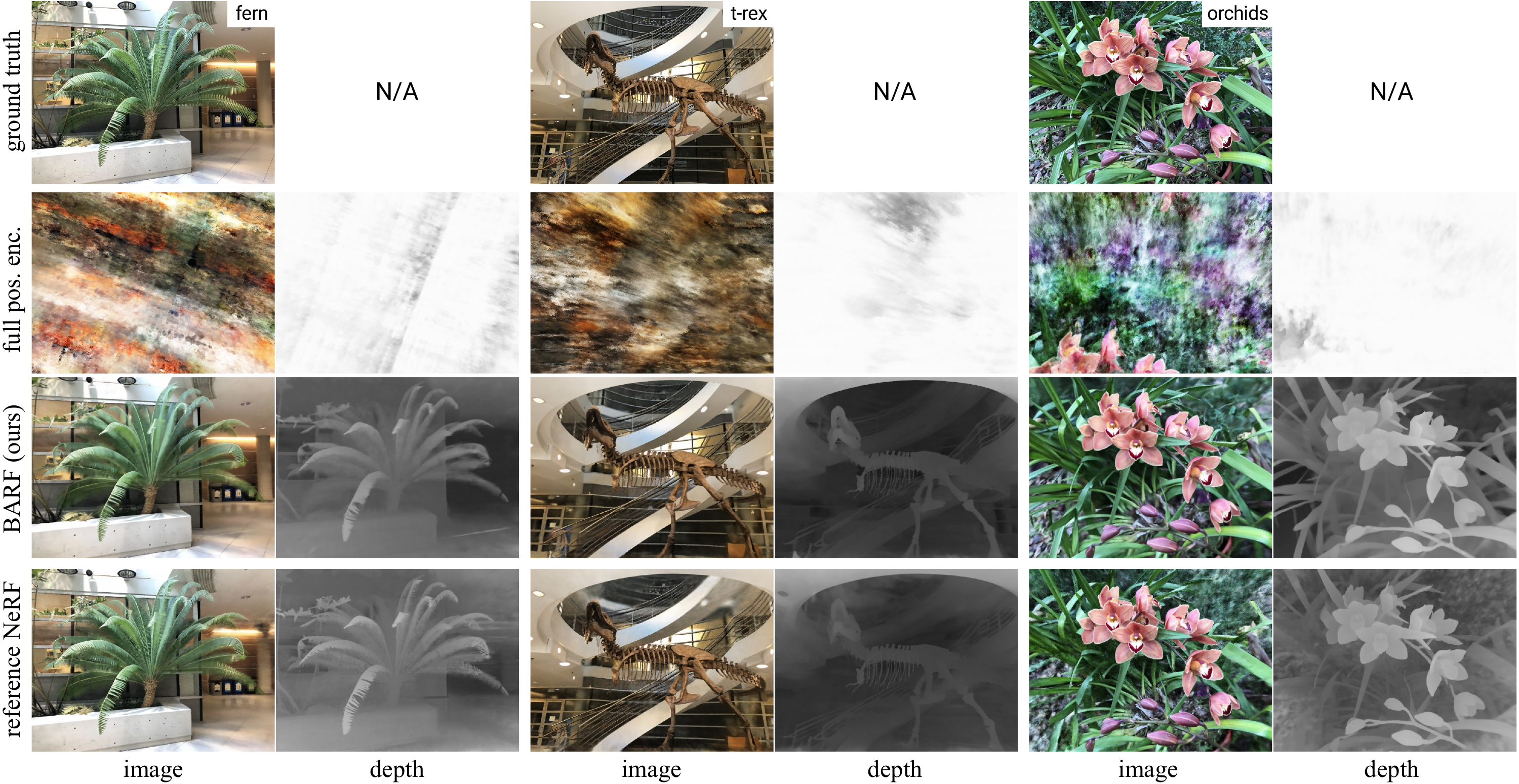}
    \vspace{-4pt}
    \caption{
        {\bf Qualitative results} of NeRF on real-world scenes from \emph{unknown} camera poses.
        Compared to a reference NeRF model trained with camera poses provided from \SFM~\cite{schonberger2016structure}, BARF can effectively optimize for the poses jointly with the scene representation.
        NeRF models with full positional encoding diverge to incorrect localization and hence poor synthesis quality.
    }
    \label{fig:nerf-results-llff}
\end{figure*}

\begin{table*}[t!]
    \centering
    \begin{minipage}{0.61\linewidth}
        \centering
        \setlength\tabcolsep{3pt}
        \resizebox{\linewidth}{!}{
            \begin{tabular}{c||cc|cc||cc|c|cc|c|cc|c}
                \toprule
                \multirow{4}{*}{Scene} & \multicolumn{4}{c||}{Camera pose registration} & \multicolumn{9}{c}{View synthesis quality} \vspace{1.5pt} \\
                & \multicolumn{2}{c|}{Rotation ($\degree$) $\downarrow$} & \multicolumn{2}{c||}{Translation $\downarrow$} & \multicolumn{3}{c|}{PSNR $\uparrow$} & \multicolumn{3}{c|}{SSIM $\uparrow$} & \multicolumn{3}{c}{LPIPS $\downarrow$} \\
                \cmidrule{2-14}
                & \small full & \small \multirow{2}{*}{BARF}
                & \small full & \small \multirow{2}{*}{BARF}
                & \small full & \small \multirow{2}{*}{BARF} & \small ref.
                & \small full & \small \multirow{2}{*}{BARF} & \small ref.
                & \small full & \small \multirow{2}{*}{BARF} & \small ref. \vspace{-2.5pt} \\
                & \small pos.enc. & \small
                & \small pos.enc. & \small
                & \small pos.enc. & \small & \small NeRF
                & \small pos.enc. & \small & \small NeRF
                & \small pos.enc. & \small & \small NeRF \\
                \midrule
                Fern       & 74.452 & \bf 0.191 & 30.167 & \bf 0.192 &  9.81 & \bf 23.79 & 23.72 & 0.187 & \bf 0.710 & 0.733 & 0.853 & \bf 0.311 & 0.262 \\
                Flower     &  2.525 & \bf 0.251 &  2.635 & \bf 0.224 & 17.08 & \bf 23.37 & 23.24 & 0.344 & \bf 0.698 & 0.668 & 0.490 & \bf 0.211 & 0.244 \\
                Fortress   & 75.094 & \bf 0.479 & 33.231 & \bf 0.364 & 12.15 & \bf 29.08 & 25.97 & 0.270 & \bf 0.823 & 0.786 & 0.807 & \bf 0.132 & 0.185 \\
                Horns      & 58.764 & \bf 0.304 & 32.664 & \bf 0.222 &  8.89 & \bf 22.78 & 20.35 & 0.158 & \bf 0.727 & 0.624 & 0.805 & \bf 0.298 & 0.421 \\
                Leaves     & 88.091 & \bf 1.272 & 13.540 & \bf 0.249 &  9.64 & \bf 18.78 & 15.33 & 0.067 & \bf 0.537 & 0.306 & 0.782 & \bf 0.353 & 0.526 \\
                Orchids    & 37.104 & \bf 0.627 & 20.312 & \bf 0.404 &  9.42 & \bf 19.45 & 17.34 & 0.085 & \bf 0.574 & 0.518 & 0.806 & \bf 0.291 & 0.307 \\
                Room       &173.811 & \bf 0.320 & 66.922 & \bf 0.270 & 10.78 & \bf 31.95 & 32.42 & 0.278 & \bf 0.940 & 0.948 & 0.871 & \bf 0.099 & 0.080 \\
                T-rex      &166.231 & \bf 1.138 & 53.309 & \bf 0.720 & 10.48 & \bf 22.55 & 22.12 & 0.158 & \bf 0.767 & 0.739 & 0.885 & \bf 0.206 & 0.244 \\
                \midrule
                Mean       & 84.509 & \bf 0.573 & 31.598 & \bf 0.331 & 11.03 & \bf 23.97 & 22.56 & 0.193 & \bf 0.722 & 0.665 & 0.787 & \bf 0.238 & 0.283 \\
                \bottomrule 
            \end{tabular}
        }
        \vspace{-2pt}
        \captionof{table}{
            {\bf Quantitative results} of NeRF on the LLFF forward-facing scenes from \emph{unknown} camera poses.
            BARF can optimize for accurate camera poses (with an average $<0.6\degree$ rotation error) and high-fidelity scene representations, enabling novel view synthesis whose quality is comparable to reference NeRF model trained under \SFM poses.
            Translation errors are scaled by $100$.
            }
        \label{table:nerf-llff}
    \end{minipage}
    \hspace{8pt}
    \begin{minipage}{0.36\linewidth}
        \includegraphics[width=1\linewidth,page=1]{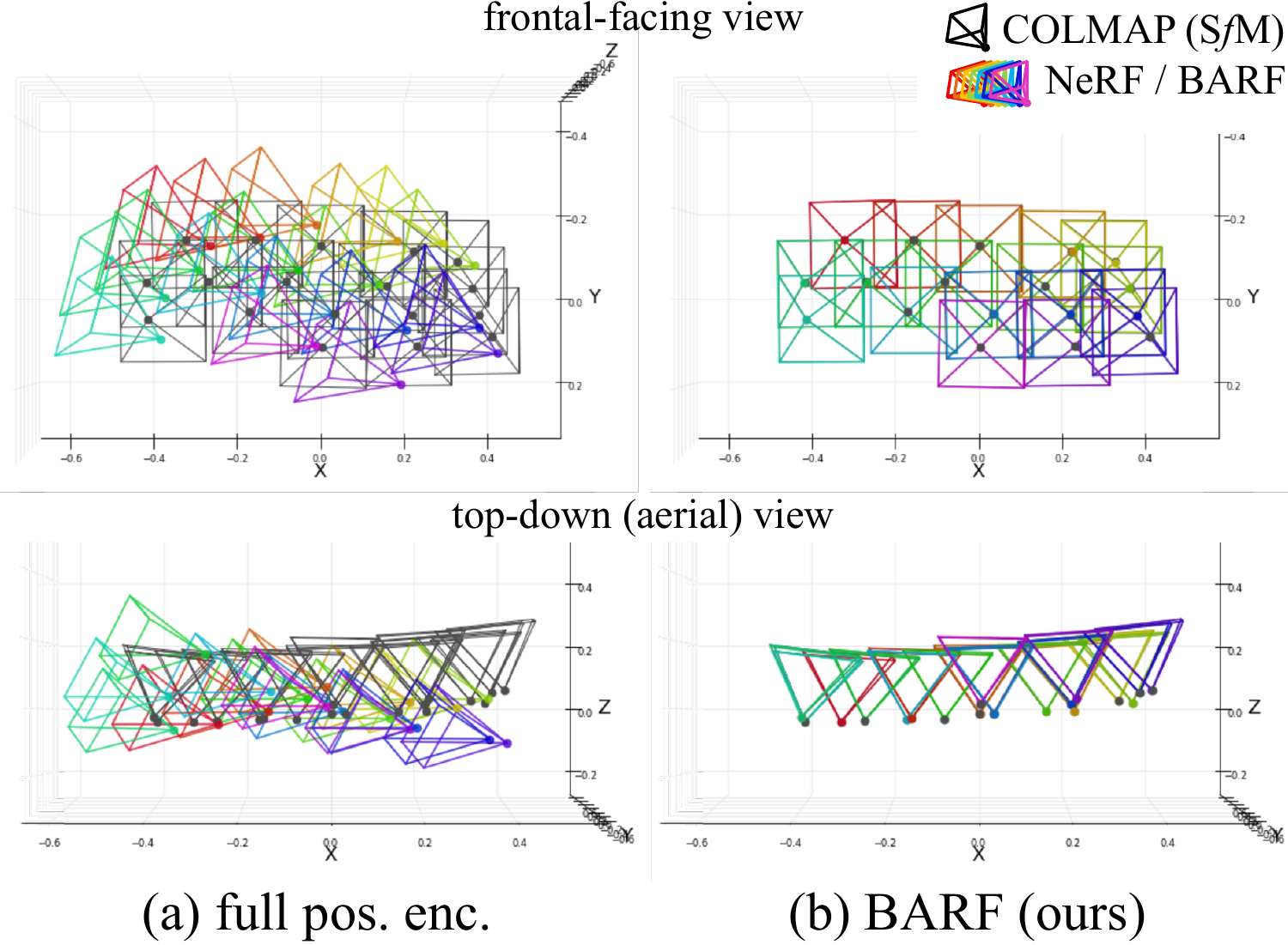}
        \vspace{-16pt}
        \captionof{figure}{
            Visualization of optimized camera poses from the {\it fern} scene (Procrustes aligned).
            Results from BARF highly agrees with \SFM, whereas the baseline poses are suboptimal.
        }
        \label{fig:nerf-llff-camera}
    \end{minipage}
\end{table*}

\vspace{1pt}
\noindent\textbf{Results.}
The quantitative results (Table~\ref{table:nerf-llff}) show that the recovered camera poses from BARF highly agrees with those estimated from off-the-shelf \SFM methods (visualized in Fig.~\ref{fig:nerf-llff-camera}), demonstrating the ability of BARF to localize from scratch.
Furthermore, BARF can successfully recover the 3D scene representation with high fidelity (Fig.~\ref{fig:nerf-results-llff}).
In contrast, NeRF with na\"ive positional encoding diverge to incorrect camera poses, which in turn results in poor view synthesis.
This highlights the effectiveness of BARF utilizing a coarse-to-fine strategy for joint registration and reconstruction.

%% file: 5-conclusion.tex

\section{Conclusion}

We present Bundle-Adjusting Neural Radiance Fields (BARF), a simple yet effective strategy for training NeRF from imperfect camera poses.
By establishing a theoretical connection to classical image alignment, we demonstrate that coarse-to-fine registration is necessary for joint registration and reconstruction with coordinate-based scene representations.
Our experiments show that BARF can effectively learn the 3D scene representations from scratch and resolve large camera pose misalignment at the same time.

Despite the intriguing results at the current stage, BARF has similar limitations to the original NeRF formulation~\cite{mildenhall2020nerf} (\eg slow optimization and rendering, rigidity assumption, sensitivity to dense 3D sampling), as well as reliance on heuristic coarse-to-fine scheduling strategies.
Nevertheless, since BARF keeps a close formulation to NeRF, many of the latest advances on improving NeRF are potentially transferable to BARF as well.
We believe BARF opens up exciting avenues for rethinking visual localization for \SFM/SLAM systems and self-supervised dense 3D reconstruction frameworks using view synthesis as a proxy objective.

\vspace{4pt}
\noindent\textbf{Acknowledgements.}
We thank Chaoyang Wang, Mengtian Li, Yen-Chen Lin, Tongzhou Wang, Sivabalan Manivasagam, and Shenlong Wang for helpful discussions and feedback on the paper.
This work was supported by the CMU Argo AI Center for Autonomous Vehicle Research.

%% file: A-supplementary.tex


\section{Visualizing the Basin of Attraction} \label{sec:supp-basin}

\begin{figure*}[t!]
    \centering  
    \includegraphics[width=1\linewidth,page=1]{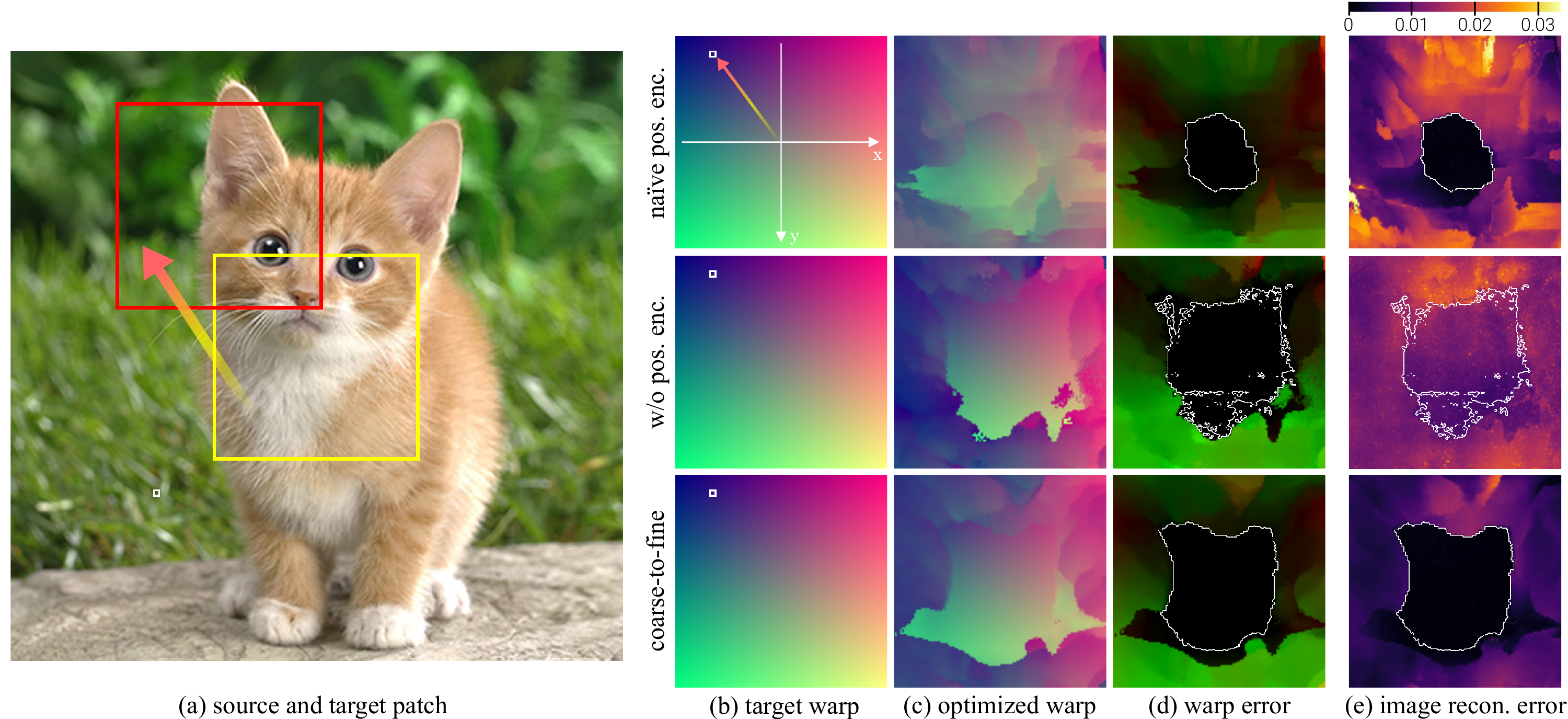}
    \caption{
        Visualization of the {\bf basin of attraction}.
        (a) We aim to align a center box (yellow) to a target patch (red) at \emph{every} possible location within the raw image.
        For each target patch, we jointly optimize $f$ and the translational warp $\p$ to analyze the final warp error and the image reconstruction loss.
        (b) The target offsets forms a color-coded map, where {\color{Green} green} indicates horizontal offsets and {\color{RubineRed} red} indicates vertical offsets.
        The above example corresponds to the highlighted pixel.
        (c) The optimized warp parameters and (d) the warp error for every target patch location, where the white contours highlight the offset error threshold of $0.5$ pixels.
        BARF effectively widens the basin of attraction (range of successful alignment) with a smoother landscape compared to na\"ive positional encoding.
        (e) Without positional encoding, $f$ has limited capacity of representing the image details, resulting in nonzero image errors despite the registration being successful as well.
    }
    \label{fig:planar-basin}
\end{figure*}

The planar image alignment setting allows us to analyze how positional encoding affects the basin of attraction.
We use the same image in Fig.~\ref{fig:planar} and consider the simpler case of aligning two image patches differing by an offset.
We use a translational warp $\p \in \Real^2$ on a square box whose size is $1/3$ of the raw image height and initialized to the raw center.
We aim to register the center box to a single target patch of the same size shifted by some offset, shown in Fig.~\ref{fig:planar-basin}(a).
We optimize the image neural network $f$ with the objective in~\eqref{eq:2d_b}, where $\I_1$ is the center patch and $\I_2$ is the target patch,
and investigate the convergence behavior of translational alignment as a function of target offsets.
We search over the entire pixel grid to as far as where the target patch has no overlapping region with the initial center box.

We visualize the results in Fig.~\ref{fig:planar-basin}.
Na\"ive positional encoding results in a more nonlinear alignment landscape and a smaller basin of attraction, while not using positional encoding sacrifices the reconstruction quality due to the limited representability of the network $f$.
In contrast, BARF can widen the basin of attraction while reconstructing the image representation with high fidelity.
This also justifies the importance of coarse-to-fine registration for NeRF in the 3D case.
Please also refer to the supplementary videos for more visualizations of the basin of attraction.


\section{Additional NeRF Details \& Results} \label{sec:supp-exp}

We provide more details and results from our NeRF experiments in this section (for real-world scenes in particular).


\subsection{Evaluation Details} \label{sec:supp-exp-nerf-llff}

As mentioned in the main paper, the optimized solutions of the 3D scenes and camera poses are up to a 3D similarity transformation.
Therefore, we evaluate the quality of registration by pre-aligning the optimized poses to the reference poses, which are the ground truth poses for the synthetic objects (Sec.~\ref{sec:exp-nerf-blender}) and pose estimation computed from \SFM packages~\cite{schonberger2016structure} for the real-world scenes (Sec.~\ref{sec:exp-nerf-llff}).

\DontPrintSemicolon
\newcommand\algcommentfont[1]{\footnotesize\textcolor{MidnightBlue!80!Gray}{#1}}
\SetCommentSty{algcommentfont}
\SetKwComment{tcp}{$\triangleright$ }{}
\SetKwFunction{KwFn}{Fn}
\SetNlSty{}{}{}

\begin{algorithm}
    \SetKwInOut{Input}{Input}
    \SetKwInOut{Output}{Output}
    \SetKwProg{Func}{Function}{:}{end}
    \Func{\sc PreAlign($\{[\Rot_i,\trans_i]\}_{i=1}^{M}, \{[\widehat{\Rot}_i,\hat{\trans}_i]\}_{i=1}^{M}$)}{
        \SetKwInOut{Input}{Input}
        \SetKwInOut{Output}{Output}
        \Input{~reference poses $\{[\Rot_i,\trans_i]\}_{i=1}^{M}$, \\ ~optimized poses $\{[\widehat{\Rot}_i,\hat{\trans}_i]\}_{i=1}^{M}$ }
        \Output{~optimized poses $\{[\widehat{\Rot}'_i,\hat{\trans}'_i]\}_{i=1}^{M}$ aligned \\ ~to the reference poses}
        \BlankLine
        \For{$i = \{1,\dots,M\}$}{
            $\o_i = -\Rot_i^\top\trans_i $ \;
            $\hat{\o}_i = -\widehat{\Rot}_i^{\top}\hat{\trans}_i $ \;
        }
        $s, \hat{s}, \trans, \hat{\trans}, \Rot = \text{\sc Procrustes($\{\o_i\}_{i=1}^{M}, \{\hat{\o}_i\}_{i=1}^{M}$)} $ \;
        \For{$i = \{1,\dots,M\}$}{
            $\hat{\o}_i' = s \Rot \left( \frac{1}{\hat{s}} (\hat{\o}_i-\hat{\trans}) \right) + \trans $ \;
            $\widehat{\Rot}_i' = \widehat{\Rot}_i \Rot^\top $ \;
            $\hat{\trans}_i' = -\widehat{\Rot}_i'^\top \hat{\o}_i' $ \;
        }
        \KwRet $\{[\widehat{\Rot}'_i,\hat{\trans}'_i]\}_{i=1}^{M}$ \;
    }
    \BlankLine
    \Func{\sc Procrustes($\{\o_i\}_{i=1}^{M}, \{\hat{\o}_i\}_{i=1}^{M}$)}{
        \Input{~reference camera centers $\{\o_i\}_{i=1}^{M}$, \\ ~optimized camera centers $\{\hat{\o}_i\}_{i=1}^{M}$ }
        \Output{~scale $s$, $\hat{s}$, translation $\trans$, $\hat{\trans}$, rotation $\Rot$}
        \BlankLine
        $\trans = \frac{1}{M}\sum_{i=1}^{M} \o_i \; \in \Real^{3} $ \;
        $\hat{\trans} = \frac{1}{M}\sum_{i=1}^{M} \hat{\o}_i \; \in \Real^{3} $ \;
        $s = \sqrt{ \frac{1}{M}\sum_{i=1}^{M} \eucnorm{\o_i - \trans}^2 } \; \in \Real $ \;
        $\hat{s} = \sqrt{ \frac{1}{M}\sum_{i=1}^{M} \eucnorm{\hat{\o}_i - \hat{\trans}}^2 } \; \in \Real $ \;
        $\X = \frac{1}{s} \left( [\o_1,\dots,\o_M] - \trans\1_M^\top \right) \; \in \Real^{3\times M}$ \; 
        $\widehat{\X} = \frac{1}{\hat{s}} \left( [\hat{\o}_1,\dots,\hat{\o}_M] - \hat{\trans}\1_M^\top \right) \; \in \Real^{3\times M}$ \; 
        $\U,\S,\V^\top = \text{SVD}(\X \widehat{\X}^\top) $ \; 
        $\Rot = \U\V^\top \; \in \Real^{3\times 3} $ \;
        \If{$\det(\Rot)=-1$}{
            multiply last row of $\Rot$ by $-1$
        }
        \KwRet $s$, $\hat{s}$, $\trans$, $\hat{\trans}$, $\Rot$ \;
    }
    \caption{Pre-align camera poses for evaluation}
    \label{alg:procrustes}
\end{algorithm}

\begin{figure*}[t!]
    \centering  
    \includegraphics[width=1\linewidth,page=1]{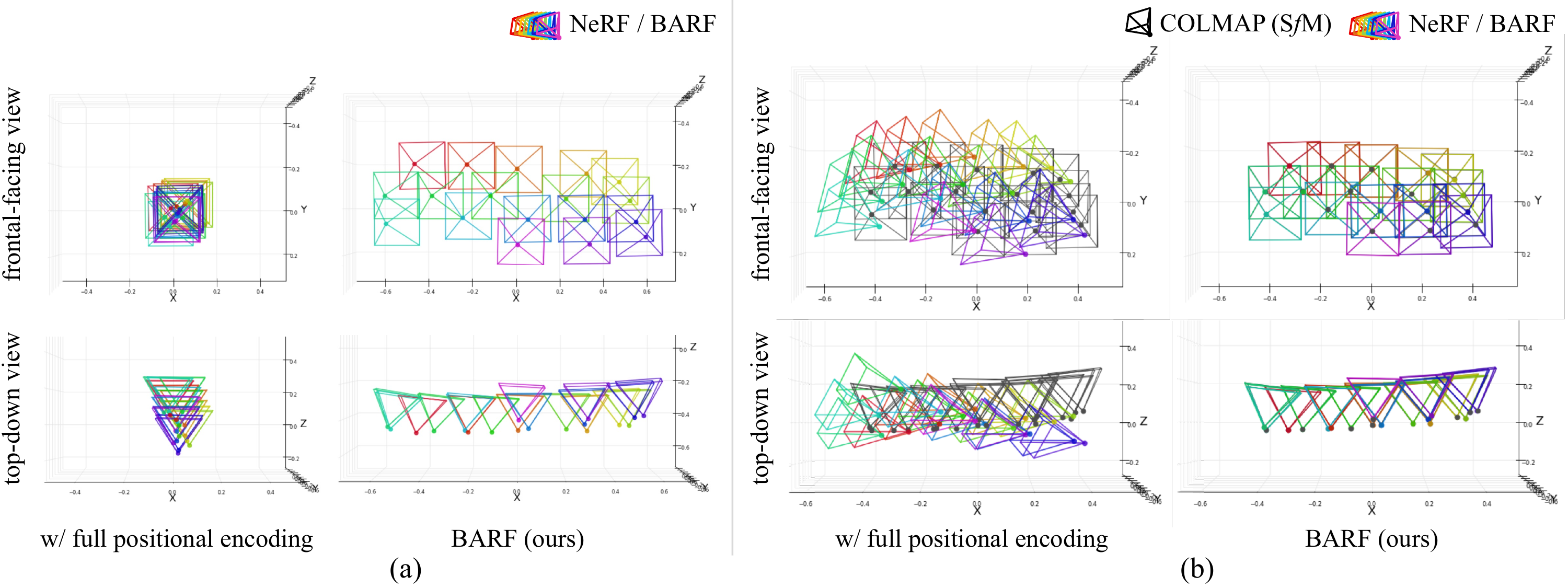}
    \caption{
        Visualization of the {\bf optimized camera poses} for the \textit{fern} scene.
        The poses for both the baseline NeRF (with full positional encoding) and BARF are initialized to the identity transform for all frames.
        (a) The camera poses of the baseline NeRF get stuck in a suboptimal solution that does not accurately reflect the actual viewpoints, whereas BARF can effectively optimize for the underlying poses.
        (b) We compare the optimized poses to those computed from \SFM~\cite{schonberger2016structure} (colored in black), where we align the pose trajectories using Procrustes analysis.
        The camera poses optimized by BARF highly agree with those from \SFM, whereas those from the baseline NeRF cannot be well-aligned with Procrustes analysis.
        Therefore, there is no systematic way of finding a reasonable set of corresponding held-out views with respect to the optimized coordinate system.
    }
    \label{fig:supp-camera}
\end{figure*}

We use Procrustes analysis on the camera locations for aligning the coordinate systems.
The algorithm details are described in Alg.~\ref{alg:procrustes}.
We write the reference poses $\{[\Rot_i,\trans_i]\}_{i=1}^{M}$ and the optimized poses $\{[\widehat{\Rot}_i,\hat{\trans}_i]\}_{i=1}^{M}$ in the form of camera extrinsic matrices, and the aligned poses can be written as $\{[\widehat{\Rot}'_i,\hat{\trans}'_i]\}_{i=1}^{M} = \text{\sc PreAlign}(\{[\Rot_i,\trans_i]\}_{i=1}^{M}, \{[\widehat{\Rot}_i,\hat{\trans}_i]\}_{i=1}^{M})$.
After the cameras are Procrustes-aligned, we apply the relative rotation (solved for via the Procrustes analysis process) to account for rotational differences.
We measure the rotation error between the \SFM poses and the aligned poses from NeRF/BARF by the angular distance as
\begin{align} \label{eq:supp-rot-error}
    \Delta \theta_i = \cos^{-1} \frac{\text{trace}\big( \Rot_i \widehat{\Rot}^{'\top}_i \big)-1}{2} \;, \;\; i = \{1,\dots,M\} \;,
\end{align}
where $\langle \cdot,\cdot \rangle$ is the quaternion inner product.
For additional clarity, we provide a more detailed visualization of the optimized camera poses in Fig.~\ref{fig:supp-camera} (for the LLFF dataset).

\begin{figure*}[t!]
    \centering  
    \includegraphics[width=0.98\linewidth,page=1]{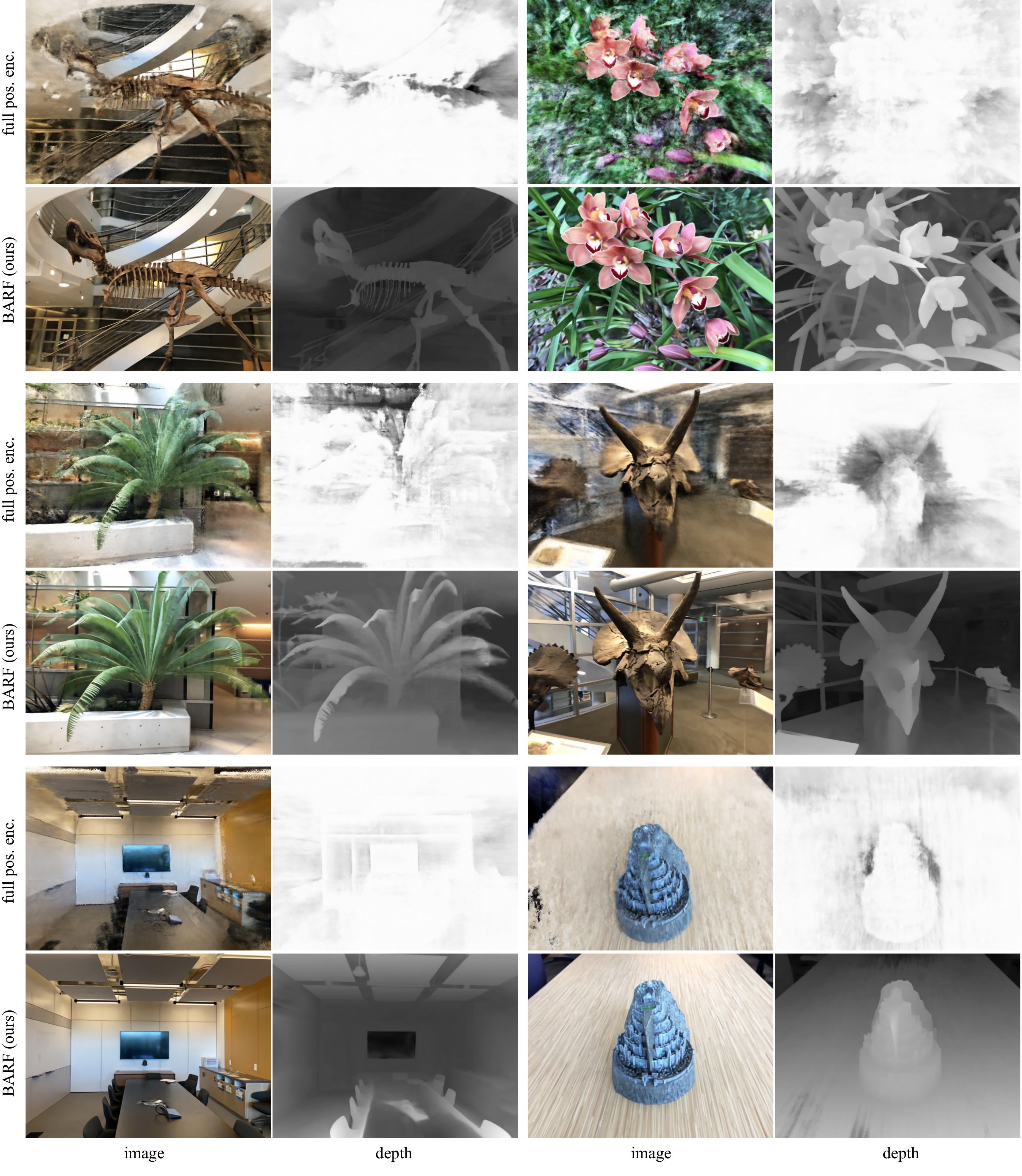}
    \caption{
    	Additional novel view synthesis results from the real-world scene experiment (LLFF dataset).
    	Instead of visualizing the held-out views computed by Procrustes analysis, we show qualitative results at new viewpoints by sampling camera pose perturbations around the viewpoint from the training set (closest to the average pose).
    	Note that for this set of qualitative results, we do not have ground-truth RGB images to compare against.
    	BARF can optimize for scene representations of much higher quality.
    	Please refer to the supplementary video for more details.
    }
    \label{fig:supp-llff}
\end{figure*}

To evaluate the quality of novel view synthesis while being minimally affected by camera misalignment, we transform the test views (provided by Mildenhall~\etal~\cite{mildenhall2019local}) to the coordinate system of the optimized poses by applying the scale/rotation/translation from the Procrustes analysis, as in Alg.~\ref{alg:procrustes}.
The camera trajectories from the baseline NeRF with na\"ive full positional encoding exhibits large rotational and translational differences compared to \SFM poses in general.
For this reason, the view synthesis results from the baseline NeRF, whose corresponding test views are also determined using Procrustes analysis, are far from plausible.
Unfortunately, there is no other systematic way of determining what the corresponding views held out from the \SFM poses would be in the learned coordinate system.
Nevertheless, we provide additional qualitative results in Fig.~\ref{fig:supp-llff}, where the novel views are selected from a \emph{training} view closest to the average pose and sampling translational perturbations.
Please also see the supplementary video for more details.


\subsection{Real-World Scenes (LLFF Dataset)} \label{sec:supp-exp-llff}

\vspace{4pt}
\noindent\textbf{Dataset.}
The LLFF dataset~\cite{mildenhall2019local} consists of 8 forward-facing scenes with RGB images sequentially captured by hand-held cameras.
In the original NeRF paper~\cite{mildenhall2019local}, the test views were selected by holding out every 8th frame from the video sequence and training with the remaining frames.
Unlike Mildenhall~\etal~\cite{mildenhall2019local}, however, we hold out the last $10\%$ of the frames for evaluation and train with the first $90\%$ frames.
This train/test split does not assume that the held-out views are interpolations of the training views, which allows a more practical simulation of predicting future viewpoints from previous observations.
The statistics of the train/test split for each scene is provided in Table~\ref{table:supp-llff-split}.

\begin{table}[t!]
    \centering
    \setlength\tabcolsep{3pt}
    \resizebox{\linewidth}{!}{
        \begin{tabular}{c|cccccccc}
            \toprule
            & Fern & Flower & Fortress & Horns & Leaves & Orchids & Room & T-rex \\
            \midrule
            split & 18\,/\,2 & 31\,/\,3 & 38\,/\,4 & 56\,/\,6 & 24\,/\,2 & 23\,/\,2 & 37\,/\,4 & 50\,/\,5 \\
            total & 20 & 34 & 42 & 62 & 26 & 25 & 41 & 55 \\
            \bottomrule 
        \end{tabular}
    }
    \caption{
        Dataset statistics of the train/test splits for the real-world scene (LLFF) experiments, where we hold out the last $10\%$ frames from each sequences.
    }
    \label{table:supp-llff-split}
\end{table}

\begin{table*}[t!]
    \centering
    \setlength\tabcolsep{3pt}
    \resizebox{\linewidth}{!}{
        \begin{tabular}{c||ccc|ccc||ccc|c|ccc|c|ccc|c}
            \toprule
            \multirow{4}{*}{Scene} & \multicolumn{6}{c||}{Camera pose registration} & \multicolumn{12}{c}{View synthesis quality} \vspace{1.5pt} \\
            & \multicolumn{3}{c|}{Rotation ($\degree$) $\downarrow$} & \multicolumn{3}{c||}{Translation $\downarrow$} & \multicolumn{4}{c|}{PSNR $\uparrow$} & \multicolumn{4}{c|}{SSIM $\uparrow$} & \multicolumn{4}{c}{LPIPS $\downarrow$} \\
            \cmidrule{2-19}
            & \small full & \small w/o & \small \multirow{2}{*}{BARF}
            & \small full & \small w/o & \small \multirow{2}{*}{BARF}
            & \small full & \small w/o & \small \multirow{2}{*}{BARF} & \small ref.
            & \small full & \small w/o & \small \multirow{2}{*}{BARF} & \small ref.
            & \small full & \small w/o & \small \multirow{2}{*}{BARF} & \small ref. \vspace{-2.5pt} \\
            & \small pos.enc. & \small pos.enc. & \small
            & \small pos.enc. & \small pos.enc. & \small
            & \small pos.enc. & \small pos.enc. & \small & \small NeRF
            & \small pos.enc. & \small pos.enc. & \small & \small NeRF
            & \small pos.enc. & \small pos.enc. & \small & \small NeRF \\
            \midrule
            Fern       & 74.452 & 0.194 & \bf 0.191 & 30.167 & 0.194 & \bf 0.192 &  9.81 & 23.73 & \bf 23.79 & 23.72 & 0.187 & 0.709 & \bf 0.710 & 0.733 & 0.853 & 0.371 & \bf 0.311 & 0.262 \\
            Flower     &  2.525 & 0.883 & \bf 0.251 &  2.635 & 0.297 & \bf 0.224 & 17.08 & \bf 24.66 & 23.37 & 23.24 & 0.344 & \bf 0.739 & 0.698 & 0.668 & 0.490 & \bf 0.200 & 0.211 & 0.244 \\
            Fortress   & 75.094 & \bf 0.320 & 0.479 & 33.231 & \bf 0.289 & 0.364 & 12.15 & 28.35 & \bf 29.08 & 25.97 & 0.270 & 0.774 & \bf 0.823 & 0.786 & 0.807 & 0.206 & \bf 0.132 & 0.185 \\
            Horns      & 58.764 & \bf 0.182 & 0.304 & 32.664 & \bf 0.170 & 0.222 &  8.89 & 22.27 & \bf 22.78 & 20.35 & 0.158 & 0.724 & \bf 0.727 & 0.624 & 0.805 & 0.312 & \bf 0.298 & 0.421 \\
            Leaves     & 88.091 & 2.938 & \bf 1.272 & 13.540 & 0.468 & \bf 0.249 &  9.64 & \bf 19.08 & 18.78 & 15.33 & 0.067 & \bf 0.566 & 0.537 & 0.306 & 0.782 & 0.375 & \bf 0.353 & 0.526 \\
            Orchids    & 37.104 & \bf 0.550 & 0.627 & 20.312 & \bf 0.396 & 0.404 &  9.42 & 19.27 & \bf 19.45 & 17.34 & 0.085 & 0.566 & \bf 0.574 & 0.518 & 0.806 & 0.313 & \bf 0.291 & 0.307 \\
            Room       &173.811 & 0.384 & \bf 0.320 & 66.922 & 0.311 & \bf 0.270 & 10.78 & 30.71 & \bf 31.95 & 32.42 & 0.278 & 0.928 & \bf 0.940 & 0.948 & 0.871 & 0.135 & \bf 0.099 & 0.080 \\
            T-rex      &166.231 & \bf 0.138 & 1.138 & 53.309 & \bf 0.261 & 0.720 & 10.48 & 22.48 & \bf 22.55 & 22.12 & 0.158 & \bf 0.783 & 0.767 & 0.739 & 0.885 & \bf 0.197 & 0.206 & 0.244 \\
            \midrule
            Mean       & 84.509 & 0.699 & \bf 0.573 & 31.598 & \bf 0.298 & 0.331 & 11.03 & 23.82 & \bf 23.97 & 22.56 & 0.193 & \bf 0.724 & 0.722 & 0.665 & 0.787 & 0.264 & \bf 0.238 & 0.283 \\
            \bottomrule 
        \end{tabular}
    }
    \caption{
        Full quantitative comparison of NeRF on the LLFF forward-facing scenes from \emph{unknown} camera poses.
        BARF and our baseline without positional encoding are competitive in different metrics.
        An optimal coarse-to-fine schedule for BARF could be theoretically found per scene that are at least as good as the baseline methods; exhaustively or adaptively search for such optimal schedule is currently out of scope of this paper.
        Translation errors are scaled by $100$.
    }
    \label{table:supp-llff-full}
    \vspace{-4pt}
\end{table*}

\vspace{4pt}
\noindent\textbf{Full comparison.}
We provide a more complete evaluation of the LLFF experiment in Table~\ref{table:supp-llff-full}, where we also include the baseline without any positional encoding.
Note that we consider the same schedule for all scenes in the dataset (adjusting the positional encoding from iterations $20$K to $100$K); due to the per-scene optimization nature, however, the optimal coarse-to-fine scheduling for each scene would actually be data-dependent.
Despite this, the coarse-to-fine scheduling considered here already allows BARF to achieve an averaged similar or better performance on real-world scenes.
An exhaustive analysis of searching for the best scheduling is currently out of scope of this paper.

\begin{table*}[t!]
    \centering
    \setlength\tabcolsep{3pt}
    \resizebox{\linewidth}{!}{
        \begin{tabular}{c||ccc|ccc||ccc|c|ccc|c|ccc|c}
            \toprule
            \multirow{4}{*}{Scene} & \multicolumn{6}{c||}{Camera pose registration} & \multicolumn{12}{c}{View synthesis quality} \vspace{1.5pt} \\
            & \multicolumn{3}{c|}{Rotation ($\degree$) $\downarrow$} & \multicolumn{3}{c||}{Translation $\downarrow$} & \multicolumn{4}{c|}{PSNR $\uparrow$} & \multicolumn{4}{c|}{SSIM $\uparrow$} & \multicolumn{4}{c}{LPIPS $\downarrow$} \\
            \cmidrule{2-19}
            & \small full & \small w/o & \small \multirow{2}{*}{BARF}
            & \small full & \small w/o & \small \multirow{2}{*}{BARF}
            & \small full & \small w/o & \small \multirow{2}{*}{BARF} & \small ref.
            & \small full & \small w/o & \small \multirow{2}{*}{BARF} & \small ref.
            & \small full & \small w/o & \small \multirow{2}{*}{BARF} & \small ref. \vspace{-2.5pt} \\
            & \small pos.enc. & \small pos.enc. & \small
            & \small pos.enc. & \small pos.enc. & \small
            & \small pos.enc. & \small pos.enc. & \small & \small NeRF
            & \small pos.enc. & \small pos.enc. & \small & \small NeRF
            & \small pos.enc. & \small pos.enc. & \small & \small NeRF \\
            \midrule
            Fern       &164.243 & \bf 0.391 & 0.448 & 18.265 & \bf 0.260 & 0.283 &  9.17 & 23.39 & \bf 23.55 & 22.76 & 0.148 & \bf 0.700 & \bf 0.700 & 0.655 & 1.041 & 0.362 & \bf 0.335 & 0.397 \\
            Flower     &  7.462 & \bf 0.177 & 3.282 &  1.959 & \bf 0.211 & 0.724 & 18.81 & \bf 23.63 & 22.99 & 23.37 & 0.408 & \bf 0.710 & 0.651 & 0.654 & 0.657 & \bf 0.224 & 0.227 & 0.272 \\
            Fortress   &172.581 & \bf 0.502 & 0.576 & 46.673 & \bf 0.466 & 0.468 & 11.17 & 26.75 & \bf 26.92 & 25.67 & 0.222 & 0.684 & \bf 0.716 & 0.662 & 1.122 & 0.348 & \bf 0.270 & 0.403 \\
            Horns      & 34.840 & \bf 0.248 & 0.266 & 18.207 & \bf 0.223 & 0.228 &  8.95 & 21.52 & \bf 21.79 & 20.37 & 0.174 & \bf 0.714 & 0.701 & 0.599 & 1.028 & 0.325 & \bf 0.310 & 0.464 \\
            Leaves     &  4.708 & \bf 1.194 & 1.832 &  1.105 & \bf 0.261 & 0.367 & 11.66 & \bf 18.36 & 17.68 & 16.34 & 0.104 & \bf 0.516 & 0.473 & 0.353 & 0.822 & 0.407 & \bf 0.356 & 0.534 \\
            Orchids    &172.600 & 0.531 & \bf 0.443 & 37.887 & \bf 0.413 & \bf 0.413 &  8.22 & \bf 18.84 & 18.57 & 16.97 & 0.062 & \bf 0.536 & 0.513 & 0.402 & 1.086 & \bf 0.357 & 0.373 & 0.564 \\
            Room       &160.757 & 0.456 & \bf 0.207 & 51.988 & 0.454 & \bf 0.203 &  8.09 & 30.90 & \bf 31.99 & 32.10 & 0.127 & 0.924 & \bf 0.938 & 0.935 & 1.215 & 0.139 & \bf 0.104 & 0.109 \\
            T-rex      &175.893 & \bf 0.334 & 5.586 & 61.026 & \bf 0.328 & 3.085 &  8.30 & \bf 22.74 & 21.24 & 22.42 & 0.123 & \bf 0.794 & 0.731 & 0.770 & 1.174 & \bf 0.187 & 0.225 & 0.205 \\
            \midrule
            Mean       &111.635 & \bf 0.479 & 1.580 & 29.639 & \bf 0.327 & 0.721 & 10.54 & \bf 23.26 & 23.09 & 22.50 & 0.171 & \bf 0.698 & 0.678 & 0.629 & 1.018 & 0.294 & \bf 0.275 & 0.368 \\
            \bottomrule 
        \end{tabular}
    }
    \caption{
        Quantitative results of NeRF on the LLFF forward-facing scenes from \emph{unknown} camera poses, sampling the 3D points in the regular depth space (instead of the inverse depth space).
        Translation errors are scaled by $100$.
    }
    \label{table:supp-llff-metric}
    \vspace{-4pt}
\end{table*}

In the main LLFF experiments, we sample 3D points along each ray linearly in the inverse depth (disparity) space, where the lower and upper bounds are the image plane and infinity respectively (\ie $1/z_\text{near}=1$ and $1/z_\text{far}=0$).
To analyze the effect of depth parametrization on the performance of real-world scenes, we run an additional set of the same experiments by sampling the 3D points in the regular (metric) depth space, bounded by $z_\text{near}=1$ and $z_\text{far}=20$.

We report the quantitative results in Table~\ref{table:supp-llff-metric}.
The baseline NeRF with full positional encoding still performs poorly in all metrics.
Although the baseline without positional encoding may be slightly better than BARF in this setup, all methods being compared here exhibit better performance when the 3D points are sampled in the inverse depth space.
We present empirical results as a supplement and leave a complete analysis of depth parametrization to future work.

%% file: main.bbl
\begin{thebibliography}{10}\itemsep=-1pt

\bibitem{agarwal2011building}
Sameer Agarwal, Yasutaka Furukawa, Noah Snavely, Ian Simon, Brian Curless,
  Steven~M Seitz, and Richard Szeliski.
\newblock Building rome in a day.
\newblock {\em ACM Communications}, 2011.

\bibitem{alismail2016photometric}
Hatem Alismail, Brett Browning, and Simon Lucey.
\newblock Photometric bundle adjustment for vision-based slam.
\newblock In {\em Asian Conference on Computer Vision}, pages 324--341.
  Springer, 2016.

\bibitem{boss2020nerd}
Mark Boss, Raphael Braun, Varun Jampani, Jonathan~T Barron, Ce Liu, and Hendrik
  Lensch.
\newblock Nerd: Neural reflectance decomposition from image collections.
\newblock {\em arXiv}, 2020.

\bibitem{chaurasia2013depth}
Gaurav Chaurasia, Sylvain Duchene, Olga Sorkine-Hornung, and George Drettakis.
\newblock Depth synthesis and local warps for plausible image-based navigation.
\newblock {\em TOG}, 2013.

\bibitem{chen1993view}
Shenchang~Eric Chen and Lance Williams.
\newblock View interpolation for image synthesis.
\newblock In {\em Proceedings of the 20th annual conference on Computer
  graphics and interactive techniques}, 1993.

\bibitem{davison2007monoslam}
Andrew~J Davison, Ian~D Reid, Nicholas~D Molton, and Olivier Stasse.
\newblock Monoslam: Real-time single camera slam.
\newblock {\em TPAMI}, 2007.

\bibitem{debevec1996modeling}
Paul~E Debevec, Camillo~J Taylor, and Jitendra Malik.
\newblock Modeling and rendering architecture from photographs: A hybrid
  geometry-and image-based approach.
\newblock In {\em Proceedings of the 23rd annual conference on Computer
  graphics and interactive techniques}, 1996.

\bibitem{delaunoy2014photometric}
Ama{\"e}l Delaunoy and Marc Pollefeys.
\newblock Photometric bundle adjustment for dense multi-view 3d modeling.
\newblock In {\em Proceedings of the IEEE Conference on Computer Vision and
  Pattern Recognition}, pages 1486--1493, 2014.

\bibitem{deng2009imagenet}
Jia Deng, Wei Dong, Richard Socher, Li-Jia Li, Kai Li, and Li Fei-Fei.
\newblock Imagenet: A large-scale hierarchical image database.
\newblock In {\em 2009 IEEE conference on computer vision and pattern
  recognition}, pages 248--255. Ieee, 2009.

\bibitem{detone2018superpoint}
Daniel DeTone, Tomasz Malisiewicz, and Andrew Rabinovich.
\newblock Superpoint: Self-supervised interest point detection and description.
\newblock In {\em CVPR}, 2018.

\bibitem{dusmanu2019d2}
Mihai Dusmanu, Ignacio Rocco, Tomas Pajdla, Marc Pollefeys, Josef Sivic,
  Akihiko Torii, and Torsten Sattler.
\newblock D2-net: A trainable cnn for joint detection and description of local
  features.
\newblock {\em arXiv}, 2019.

\bibitem{engel2017direct}
Jakob Engel, Vladlen Koltun, and Daniel Cremers.
\newblock Direct sparse odometry.
\newblock {\em TPAMI}, 2017.

\bibitem{engel2014lsd}
Jakob Engel, Thomas Sch{\"o}ps, and Daniel Cremers.
\newblock Lsd-slam: Large-scale direct monocular slam.
\newblock In {\em European conference on computer vision}, pages 834--849.
  Springer, 2014.

\bibitem{eslami2018neural}
SM~Ali Eslami, Danilo~Jimenez Rezende, Frederic Besse, Fabio Viola, Ari~S
  Morcos, Marta Garnelo, Avraham Ruderman, Andrei~A Rusu, Ivo Danihelka, Karol
  Gregor, et~al.
\newblock Neural scene representation and rendering.
\newblock {\em Science}, 360(6394):1204--1210, 2018.

\bibitem{flynn2019deepview}
John Flynn, Michael Broxton, Paul Debevec, Matthew DuVall, Graham Fyffe, Ryan
  Overbeck, Noah Snavely, and Richard Tucker.
\newblock Deepview: View synthesis with learned gradient descent.
\newblock In {\em Proceedings of the IEEE Conference on Computer Vision and
  Pattern Recognition}, pages 2367--2376, 2019.

\bibitem{gao2020portrait}
Chen Gao, Yichang Shih, Wei-Sheng Lai, Chia-Kai Liang, and Jia-Bin Huang.
\newblock Portrait neural radiance fields from a single image.
\newblock {\em arXiv}, 2020.

\bibitem{hartley2004}
Richard Hartley and Andrew Zisserman.
\newblock {\em Multiple View Geometry in Computer Vision}.
\newblock Cambridge University Press, ISBN: 0521540518, second edition, 2004.

\bibitem{hedman2017casual}
Peter Hedman, Suhib Alsisan, Richard Szeliski, and Johannes Kopf.
\newblock Casual 3d photography.
\newblock {\em TOG}, 2017.

\bibitem{heigl1999plenoptic}
Benno Heigl, Reinhard Koch, Marc Pollefeys, Joachim Denzler, and Luc Van~Gool.
\newblock Plenoptic modeling and rendering from image sequences taken by a
  hand-held camera.
\newblock 1999.

\bibitem{hu2020worldsheet}
Ronghang Hu and Deepak Pathak.
\newblock Worldsheet: Wrapping the world in a 3d sheet for view synthesis from
  a single image.
\newblock {\em arXiv}, 2020.

\bibitem{kingma2014adam}
Diederik Kingma and Jimmy Ba.
\newblock Adam: A method for stochastic optimization.
\newblock In {\em International Conference on Learning Representations}, 2015.

\bibitem{kopf2014first}
Johannes Kopf, Michael~F Cohen, and Richard Szeliski.
\newblock First-person hyper-lapse videos.
\newblock {\em TOG}, 2014.

\bibitem{levoy1990efficient}
Marc Levoy.
\newblock Efficient ray tracing of volume data.
\newblock {\em ACM Transactions on Graphics (TOG)}, 9(3):245--261, 1990.

\bibitem{levoy1996light}
Marc Levoy and Pat Hanrahan.
\newblock Light field rendering.
\newblock In {\em Proceedings of the 23rd annual conference on Computer
  graphics and interactive techniques}, 1996.

\bibitem{li2020neural}
Zhengqi Li, Simon Niklaus, Noah Snavely, and Oliver Wang.
\newblock Neural scene flow fields for space-time view synthesis of dynamic
  scenes.
\newblock {\em arXiv}, 2020.

\bibitem{lin2019photometric}
Chen-Hsuan Lin, Oliver Wang, Bryan~C Russell, Eli Shechtman, Vladimir~G Kim,
  Matthew Fisher, and Simon Lucey.
\newblock Photometric mesh optimization for video-aligned 3d object
  reconstruction.
\newblock In {\em IEEE Conference on Computer Vision and Pattern Recognition
  ({CVPR})}, 2019.

\bibitem{lombardi2019neural}
Stephen Lombardi, Tomas Simon, Jason Saragih, Gabriel Schwartz, Andreas
  Lehrmann, and Yaser Sheikh.
\newblock Neural volumes: Learning dynamic renderable volumes from images.
\newblock {\em arXiv preprint arXiv:1906.07751}, 2019.

\bibitem{lucas1981iterative}
Bruce~D. Lucas and Takeo Kanade.
\newblock An iterative image registration technique with an application to
  stereo vision.
\newblock In {\em Proceedings of the 7th International Joint Conference on
  Artificial Intelligence - Volume 2}, IJCAI'81, pages 674--679, 1981.

\bibitem{meshry2019neural}
Moustafa Meshry, Dan~B Goldman, Sameh Khamis, Hugues Hoppe, Rohit Pandey, Noah
  Snavely, and Ricardo Martin-Brualla.
\newblock Neural rerendering in the wild.
\newblock In {\em Proceedings of the IEEE Conference on Computer Vision and
  Pattern Recognition}, pages 6878--6887, 2019.

\bibitem{mildenhall2019local}
Ben Mildenhall, Pratul~P Srinivasan, Rodrigo Ortiz-Cayon, Nima~Khademi
  Kalantari, Ravi Ramamoorthi, Ren Ng, and Abhishek Kar.
\newblock Local light field fusion: Practical view synthesis with prescriptive
  sampling guidelines.
\newblock {\em ACM Transactions on Graphics (TOG)}, 38(4):1--14, 2019.

\bibitem{mildenhall2020nerf}
Ben Mildenhall, Pratul~P Srinivasan, Matthew Tancik, Jonathan~T Barron, Ravi
  Ramamoorthi, and Ren Ng.
\newblock Nerf: Representing scenes as neural radiance fields for view
  synthesis.
\newblock In {\em European conference on computer vision}, 2020.

\bibitem{mur2015orb}
Raul Mur-Artal, Jose Maria~Martinez Montiel, and Juan~D Tardos.
\newblock Orb-slam: a versatile and accurate monocular slam system.
\newblock {\em IEEE transactions on robotics}, 31(5):1147--1163, 2015.

\bibitem{newcombe2011dtam}
Richard~A Newcombe, Steven~J Lovegrove, and Andrew~J Davison.
\newblock Dtam: Dense tracking and mapping in real-time.
\newblock In {\em ICCV}, 2011.

\bibitem{niemeyer2020giraffe}
Michael Niemeyer and Andreas Geiger.
\newblock Giraffe: Representing scenes as compositional generative neural
  feature fields.
\newblock {\em arXiv}, 2020.

\bibitem{ono2018lf}
Yuki Ono, Eduard Trulls, Pascal Fua, and Kwang~Moo Yi.
\newblock Lf-net: Learning local features from images.
\newblock {\em arXiv}, 2018.

\bibitem{park2020deformable}
Keunhong Park, Utkarsh Sinha, Jonathan~T Barron, Sofien Bouaziz, Dan~B Goldman,
  Steven~M Seitz, and Ricardo-Martin Brualla.
\newblock Deformable neural radiance fields.
\newblock {\em arXiv preprint arXiv:2011.12948}, 2020.

\bibitem{pollefeys1999self}
Marc Pollefeys, Reinhard Koch, and Luc Van~Gool.
\newblock Self-calibration and metric reconstruction inspite of varying and
  unknown intrinsic camera parameters.
\newblock {\em International Journal of Computer Vision}, 32(1):7--25, 1999.

\bibitem{pollefeys2004visual}
Marc Pollefeys, Luc Van~Gool, Maarten Vergauwen, Frank Verbiest, Kurt Cornelis,
  Jan Tops, and Reinhard Koch.
\newblock Visual modeling with a hand-held camera.
\newblock {\em International Journal of Computer Vision}, 59(3):207--232, 2004.

\bibitem{pumarola2020d}
Albert Pumarola, Enric Corona, Gerard Pons-Moll, and Francesc Moreno-Noguer.
\newblock D-nerf: Neural radiance fields for dynamic scenes.
\newblock {\em arXiv}, 2020.

\bibitem{rebain2020derf}
Daniel Rebain, Wei Jiang, Soroosh Yazdani, Ke Li, Kwang~Moo Yi, and Andrea
  Tagliasacchi.
\newblock Derf: Decomposed radiance fields.
\newblock {\em arXiv}, 2020.

\bibitem{rematas2021sharf}
Konstantinos Rematas, Ricardo Martin-Brualla, and Vittorio Ferrari.
\newblock Sharf: Shape-conditioned radiance fields from a single view.
\newblock {\em arXiv}, 2021.

\bibitem{riegler2020free}
Gernot Riegler and Vladlen Koltun.
\newblock Free view synthesis.
\newblock In {\em ECCV}, 2020.

\bibitem{riegler2020stable}
Gernot Riegler and Vladlen Koltun.
\newblock Stable view synthesis.
\newblock {\em arXiv}, 2020.

\bibitem{schonberger2016structure}
Johannes~L Schonberger and Jan-Michael Frahm.
\newblock Structure-from-motion revisited.
\newblock In {\em Proceedings of the IEEE conference on computer vision and
  pattern recognition}, pages 4104--4113, 2016.

\bibitem{schwarz2020graf}
Katja Schwarz, Yiyi Liao, Michael Niemeyer, and Andreas Geiger.
\newblock Graf: Generative radiance fields for 3d-aware image synthesis.
\newblock {\em arXiv}, 2020.

\bibitem{shih20203d}
Meng-Li Shih, Shih-Yang Su, Johannes Kopf, and Jia-Bin Huang.
\newblock 3d photography using context-aware layered depth inpainting.
\newblock In {\em CVPR}, 2020.

\bibitem{sitzmann2019scene}
Vincent Sitzmann, Michael Zollh{\"o}fer, and Gordon Wetzstein.
\newblock Scene representation networks: Continuous 3d-structure-aware neural
  scene representations.
\newblock In {\em Advances in Neural Information Processing Systems}, 2016.

\bibitem{snavely2006photo}
Noah Snavely, Steven~M Seitz, and Richard Szeliski.
\newblock Photo tourism: exploring photo collections in 3d.
\newblock In {\em SIGGRAPH}. 2006.

\bibitem{snavely2008modeling}
Noah Snavely, Steven~M Seitz, and Richard Szeliski.
\newblock Modeling the world from internet photo collections.
\newblock {\em IJCV}, 2008.

\bibitem{srinivasan2020nerv}
Pratul~P Srinivasan, Boyang Deng, Xiuming Zhang, Matthew Tancik, Ben
  Mildenhall, and Jonathan~T Barron.
\newblock Nerv: Neural reflectance and visibility fields for relighting and
  view synthesis.
\newblock {\em arXiv}, 2020.

\bibitem{srinivasan2019pushing}
Pratul~P Srinivasan, Richard Tucker, Jonathan~T Barron, Ravi Ramamoorthi, Ren
  Ng, and Noah Snavely.
\newblock Pushing the boundaries of view extrapolation with multiplane images.
\newblock In {\em Proceedings of the IEEE Conference on Computer Vision and
  Pattern Recognition}, pages 175--184, 2019.

\bibitem{szeliski1998stereo}
Richard Szeliski and Polina Golland.
\newblock Stereo matching with transparency and matting.
\newblock In {\em ICCV}, 1998.

\bibitem{tancik2020fourier}
Matthew Tancik, Pratul~P Srinivasan, Ben Mildenhall, Sara Fridovich-Keil,
  Nithin Raghavan, Utkarsh Singhal, Ravi Ramamoorthi, Jonathan~T Barron, and
  Ren Ng.
\newblock Fourier features let networks learn high frequency functions in low
  dimensional domains.
\newblock In {\em Advances in Neural Information Processing Systems}, 2020.

\bibitem{tewari2020state}
Ayush Tewari, Ohad Fried, Justus Thies, Vincent Sitzmann, Stephen Lombardi,
  Kalyan Sunkavalli, Ricardo Martin-Brualla, Tomas Simon, Jason Saragih,
  Matthias Nie{\ss}ner, et~al.
\newblock State of the art on neural rendering.
\newblock In {\em Computer Graphics Forum}, volume~39, pages 701--727. Wiley
  Online Library, 2020.

\bibitem{tucker2020single}
Richard Tucker and Noah Snavely.
\newblock Single-view view synthesis with multiplane images.
\newblock In {\em Proceedings of the IEEE/CVF Conference on Computer Vision and
  Pattern Recognition}, pages 551--560, 2020.

\bibitem{tulsiani2018layer}
Shubham Tulsiani, Richard Tucker, and Noah Snavely.
\newblock Layer-structured 3d scene inference via view synthesis.
\newblock In {\em ECCV}, 2018.

\bibitem{vaswani2017attention}
Ashish Vaswani, Noam Shazeer, Niki Parmar, Jakob Uszkoreit, Llion Jones,
  Aidan~N Gomez, {\L}ukasz Kaiser, and Illia Polosukhin.
\newblock Attention is all you need.
\newblock {\em Advances in Neural Information Processing Systems},
  30:5998--6008, 2017.

\bibitem{wang2018learning}
Chaoyang Wang, Jos{\'e}~Miguel Buenaposada, Rui Zhu, and Simon Lucey.
\newblock Learning depth from monocular videos using direct methods.
\newblock In {\em CVPR}, 2018.

\bibitem{wang2017stereo}
Rui Wang, Martin Schworer, and Daniel Cremers.
\newblock Stereo dso: Large-scale direct sparse visual odometry with stereo
  cameras.
\newblock In {\em ICCV}, 2017.

\bibitem{wang2021nerf}
Zirui Wang, Shangzhe Wu, Weidi Xie, Min Chen, and Victor~Adrian Prisacariu.
\newblock Nerf $--$: Neural radiance fields without known camera parameters.
\newblock {\em arXiv}, 2021.

\bibitem{wiles2020synsin}
Olivia Wiles, Georgia Gkioxari, Richard Szeliski, and Justin Johnson.
\newblock Synsin: End-to-end view synthesis from a single image.
\newblock In {\em Proceedings of the IEEE/CVF Conference on Computer Vision and
  Pattern Recognition}, pages 7467--7477, 2020.

\bibitem{wu2011visualsfm}
Changchang Wu et~al.
\newblock Visualsfm: A visual structure from motion system.

\bibitem{xian2020space}
Wenqi Xian, Jia-Bin Huang, Johannes Kopf, and Changil Kim.
\newblock Space-time neural irradiance fields for free-viewpoint video.
\newblock {\em arXiv}, 2020.

\bibitem{yang2021asynchronous}
Anqi~Joyce Yang, Can Cui, Ioan~Andrei B{\^a}rsan, Raquel Urtasun, and Shenlong
  Wang.
\newblock Asynchronous multi-view {SLAM}.
\newblock In {\em ICRA}, 2021.

\bibitem{yen2020inerf}
Lin Yen-Chen, Pete Florence, Jonathan~T Barron, Alberto Rodriguez, Phillip
  Isola, and Tsung-Yi Lin.
\newblock inerf: Inverting neural radiance fields for pose estimation.
\newblock {\em arXiv preprint arXiv:2012.05877}, 2020.

\bibitem{yin2018geonet}
Zhichao Yin and Jianping Shi.
\newblock Geonet: Unsupervised learning of dense depth, optical flow and camera
  pose.
\newblock In {\em CVPR}, 2018.

\bibitem{yu2020pixelnerf}
Alex Yu, Vickie Ye, Matthew Tancik, and Angjoo Kanazawa.
\newblock pixelnerf: Neural radiance fields from one or few images.
\newblock In {\em CVPR}, 2021.

\bibitem{zhang2020nerf++}
Kai Zhang, Gernot Riegler, Noah Snavely, and Vladlen Koltun.
\newblock Nerf++: Analyzing and improving neural radiance fields.
\newblock {\em arXiv}, 2020.

\bibitem{zhang2018unreasonable}
Richard Zhang, Phillip Isola, Alexei~A Efros, Eli Shechtman, and Oliver Wang.
\newblock The unreasonable effectiveness of deep features as a perceptual
  metric.
\newblock In {\em Proceedings of the IEEE conference on computer vision and
  pattern recognition}, pages 586--595, 2018.

\bibitem{zhou2017unsupervised}
Tinghui Zhou, Matthew Brown, Noah Snavely, and David~G Lowe.
\newblock Unsupervised learning of depth and ego-motion from video.
\newblock In {\em CVPR}, 2017.

\bibitem{zhou2018stereo}
Tinghui Zhou, Richard Tucker, John Flynn, Graham Fyffe, and Noah Snavely.
\newblock Stereo magnification: Learning view synthesis using multiplane
  images.
\newblock In {\em SIGGRAPH}, 2018.

\bibitem{zitnick2004high}
C~Lawrence Zitnick, Sing~Bing Kang, Matthew Uyttendaele, Simon Winder, and
  Richard Szeliski.
\newblock High-quality video view interpolation using a layered representation.
\newblock {\em TOG}, 2004.

\end{thebibliography}
